\documentclass[acmtog]{acmart}
\AtBeginDocument{%
  }

\usepackage{array}
\usepackage{cleveref}
\usepackage{listings}
\usepackage{multirow}
\usepackage{placeins}
\usepackage{tikz}
\usepackage{subcaption}

\definecolor{myGreen}{RGB}{144, 238, 144}
\definecolor{myRed}{RGB}{255, 204, 203}

\newcommand{\method}{MemComposer}

\newcommand{\prefbar}[1]{%
  \begin{tikzpicture}[
      baseline={(current bounding box.center |- 0,2pt)},
      outer sep=0,
    ]
    \def\barwidth{2.7}   
    \def\barheight{0.31} 

    \pgfmathsetmacro{\redFrac}{#1/100}
    \pgfmathsetmacro{\greenFrac}{1 - \redFrac}

    \pgfmathsetmacro{\oursPct}{100 - #1}

    \draw[gray] (0,0) rectangle (\barwidth,\barheight);

    \fill[myGreen] (0,0) rectangle ({\greenFrac*\barwidth}, \barheight);

    \fill[myRed] ({\greenFrac*\barwidth},0) rectangle (\barwidth,\barheight);

    \draw[
      black,
      line width=0.3pt,
      dash pattern=on 0.3pt off 0.5pt
    ]
      ({0.5*\barwidth},0) -- ({0.5*\barwidth},\barheight);

    \node[
      font=\normalsize,
      anchor=west,
      inner sep=1pt
    ]
      at (0.02*\barwidth, 0.5*\barheight)
      {\normalsize \pgfmathprintnumber[fixed,precision=1]{\oursPct}\%};
  \end{tikzpicture}%
}

\copyrightyear{2026}
\acmYear{2026}
\setcopyright{cc}
\setcctype{by}
\acmConference[SA Conference Papers '26]{SIGGRAPH Asia 2026 Conference Papers}{December 01--04, 2026}{Kuala Lumpur, Malaysia}
\acmBooktitle{SIGGRAPH Asia 2026 Conference Papers (SA Conference Papers '26), December 01--04, 2026, Kuala Lumpur, Malaysia}
\acmDOI{10.1145/3829340.3842282}
\acmISBN{979-8-4007-2842-6/2026/12}

\acmSubmissionID{1891}

\begin{document}

\title{Memory-Guided B-Roll Generation from User Video Collections}

\author{Cusuh Ham}
\email{ham@adobe.com}
\orcid{0009-0002-2686-052X}
\affiliation{%
  \institution{Adobe Research}
  \city{Cambridge}
  \country{USA}
}

\author{Fabian Caba Heilbron}
\affiliation{%
  \institution{Adobe Research}
  \city{San Jose}
  \country{USA}}
\email{caba@adobe.com}

\author{Josef Sivic}
\affiliation{%
  \institution{Adobe Research}
  \country{USA}
}
\affiliation{%
  \institution{Czech Institute of Informatics, Robotics and Cybernetics, Czech Technical University}
  \country{Czech Republic}
}
\email{inr03127@adobe.com}

\author{Bryan Russell}
\affiliation{%
 \institution{Adobe Research}
 \city{San Francisco}
 \country{USA}}
\email{brussell@adobe.com}

\renewcommand{\shortauthors}{Ham et al.}

\begin{abstract}

We introduce an approach for collection-grounded B-roll sequence generation. Given a user's video collection, a directive given in natural language, and a target duration, the goal is to produce a multi-shot sequence that complements the user's primary footage (A-roll) while preserving the collection's characters, settings, objects, and style.
This task is challenging as one must choose the visual evidence from hours of captured footage that should guide the generation of each shot in the sequence.
We address this challenge with \method, a three-stage system that turns raw footage into a structured memory with visual references (characters, settings, objects, and style) and uses it to plan, retrieve, and generate grounded B-roll sequences.
First, in a one-time offline stage, \method~ constructs an entity-centric memory from raw video.
Second, it uses the memory and user directive to plan a grounded sequence and retrieve conditioning frames for each shot.
Third, it iteratively generates and critiques the sequence to enforce identity, setting, and sequence-level consistency.
We evaluate \method~ in a user preference study along two dimensions: prompt adherence and visual alignment to the user's collection. Against an ungrounded text-to-video planner, \method~ wins 60.0\% of prompt-adherence and 92.8\% of visual-alignment comparisons, showing the grounding benefit of collection memory and reference retrieval.
Against retrieval-only sequences assembled from captured footage, \method~ wins 94.5\% of prompt-adherence comparisons, showing the value of generating missing shots, while retrieval-only sequences are preferred for visual alignment in 58.2\% of comparisons. Project page: \url{https://cusuh.github.io/MemComposer}

\end{abstract}


\begin{teaserfigure}
  \includegraphics[width=\textwidth]{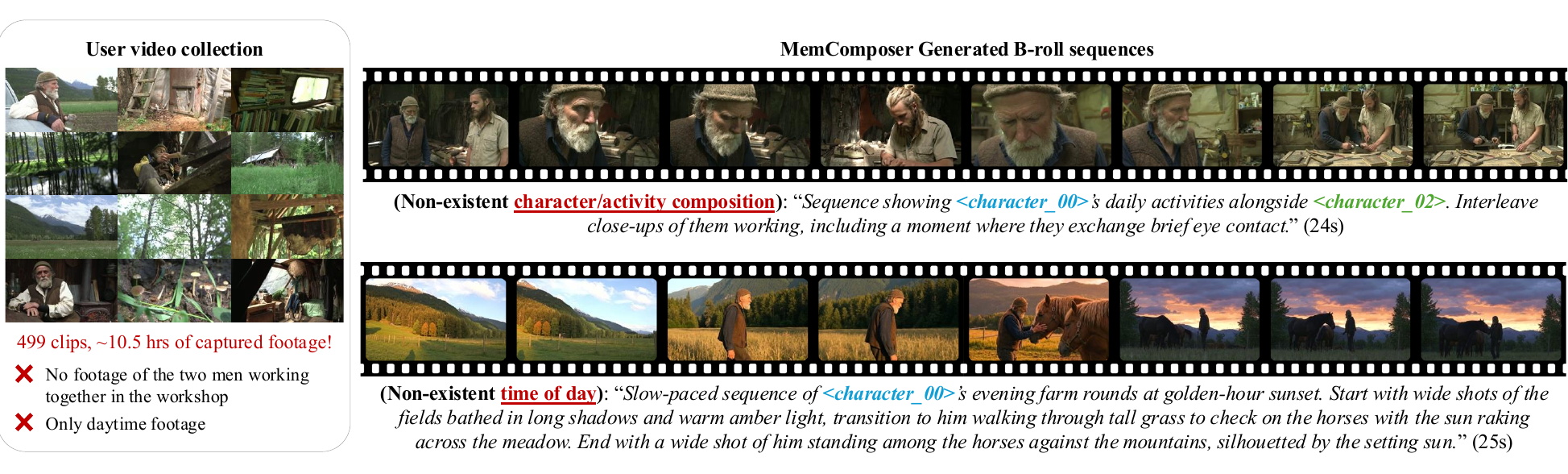}
  \caption{Given a user's video collection comprising over 10 hours of captured footage, we introduce an approach for generating a B-roll sequence that adheres to a user request given in natural language while preserving the characters, settings, and style of the captured footage. Key to our approach is the ability to construct a memory representation of the video collection for retrieving target characters and settings. {\bf Please view these results in the companion video.} Collection thumbnails are copyrighted and belong to EditStock.}
  \label{fig:teaser}
\end{teaserfigure}


\maketitle

\section{Introduction}

Picture a documentary filmmaker reviewing forty hours of interview footage. The A-roll is there: talking heads, emotional close-ups, the raw story. What's needed now is B-roll, the companion visual footage that highlights and augments the narrative: a wide shot of the neighborhood the subject grew up in, a close-up of hands turning the pages of a photo album, a sunset over the factory that closed down. Some of these shots exist somewhere in the captured footage, others were never filmed at all. Traditionally, the only options are to go back and shoot (expensive, sometimes impossible) or settle for what you have. Figure \ref{fig:teaser} shows examples of what we would like to do instead: generate new B-roll that looks like it belongs in the same visual world as the captured content.

This limitation is familiar to anyone who edits video. Content creators build compelling videos by first capturing content, then editing it into shape: trimming, sequencing, and layering A-roll and B-roll into a final piece. A lot of planning goes into the capture process precisely because of this constraint, but the framework limits artistic freedom during editing. An editor may want to show a scene from a different angle, remove visual artifacts like a shaky camera, or imagine an entirely new shot with the same characters and background.

Image and video generative models offer a way out of this bind, expanding flexibility during the editing process. However, generation should not operate independently of the captured content; the captured footage gives context to what should be generated. 
While conditioning generation models with keyframes from the collection is a powerful interface, it shifts the burden to the creator, who must manually find and curate the right references for every generated shot.
Such manual curation does not scale when collections are large (hours of footage) and creators need many missing shots.
A user should be able to identify an A-roll insertion point, give a high-level directive referencing particular characters or scenes in the captured content, and the system should know how to generate new footage in context. The generated footage should respect the overall look and storyline while faithfully preserving the specific identities of characters, the layout of scenes, and distinct scene elements.

In this work, we take on B-roll generation in context of user captured footage: given a corpus of captured footage and a high-level user directive describing the desired sequence, produce a sequence of B-roll clips that are visually grounded in the source material. The problem comes up frequently in practice: a content creator has incomplete coverage, a documentary filmmaker is working from archival footage that cannot be reshot, a video editor is assembling a memory piece from scattered home videos. In each case, the existing footage does not cover the story.

Getting this right requires solving three problems: how we represent, retrieve, and generate guided by source content. (i) The corpus must be indexed by entities, not moments. Generation needs to know that the person in clip 3 and the person in clip 27 are the same individual, what their permanent features look like from multiple angles, and which frames carry the strongest identity signal for conditioning. (ii) Retrieval must serve generation, not understanding. The right reference frame is not the most semantically similar one but the one with the right view angle, identity clarity, and level of detail to condition a generative model for a shot that does not yet exist. (iii) Even with good representation and retrieval, independently generated shots can fail as a sequence. Characters drift in appearance across shots, narrative flow breaks, and no single shot's quality score can detect these failures in isolation. Maintaining consistency requires iterative judging at multiple scales, where per-shot and sequence-level assessments drive targeted regeneration.

Prior work addresses pieces of this problem but not the whole. Retrieval-augmented generation~\cite{chen2023reimagen, yuan2025finerag, shalev2026imagerag} retrieves from generic databases without entity indexing and operates one image at a time, so identity cannot persist across shots. Multi-shot storyboard and long-form video systems~\cite{zhou2024storydiffusion, huang2026dreamshot, zhao2024moviedreamer} maintain consistency within their own generated outputs but assume small curated reference sets or build memory from synthetic frames. Identity-preserving video models~\cite{yuan2025identity, wei2024dreamvideo} generate faithful single clips but have no corpus-level memory connecting shots. B-roll editing tools~\cite{zhou2026autocut, sandoval2025editduet} assemble sequences from existing clips but cannot produce a shot that was never filmed. 

Generation guided by a video corpus, it turns out, requires a different kind of visual memory than understanding a video corpus does. Understanding needs semantic summaries while generation needs identity-preserving, appearance-anchored references that can directly condition a generative model. We need a memory that knows not just ``there is a person in this scene'' but ``this is Maria, she has short gray hair and a red scarf, and here are the frames where her face is clearly visible from multiple angles.''

We introduce \method, a system for generating B-roll sequences that are faithful to a user's high-level directive and the visual world of their footage. Our contributions are threefold:\\
\noindent\textbf{(1)} We construct a structured visual memory that indexes a video corpus by entities (characters, settings, actions, props), consolidating identity across clips and preserving the appearance detail that generation models require as conditioning.\\
\noindent\textbf{(2)} We propose a generation-informed retrieval mechanism that selects reference frames based on their effectiveness for generative model conditioning, using character-gated retrieval and view-diversity constraints to ensure identity fidelity.\\
\noindent\textbf{(3)} We design a multi-scale iterative refinement loop: per-shot judges that verify identity and setting fidelity, and a sequence-level critic that detects failures only visible in aggregate (repetitive shots in the B-roll sequence, character drift) and drives targeted regeneration.

Together, these components form an integrated system where memory, retrieval, and generation reinforce each other. We evaluate \method~against previous B-roll editing methods that operate only over existing content, using user directives that are feasible from the source footage for fair comparison. \method~generates sequences that adhere to the user directive better than previous methods (94.5\% win rate). It also produces B-roll that looks like it belongs in the same visual world as the captured content: in human evaluations, \method~is preferred over text-to-video with planning in 92.8\% of cases. Notably, when compared to an approach for assembling real captured footage, users judged \method~ more visually aligned in 41.8\% of comparisons. We also show qualitative results of \method~ in our companion video.

\section{Related Work}
\label{sec:related_work}

\paragraph{Retrieval-augmented visual generation.} Retrieval augmented generation for images conditions a diffusion model on retrieved visual references to enable generation of specific entities. 
Re-Imagen~\cite{chen2023reimagen} introduced the idea.
Subsequent work refined retrieval granularity~\cite{yuan2025finerag}, added reactivity to generation failures~\cite{shalev2026imagerag}, and personalized retrieval to user preferences~\cite{ling2026ragar}. 
These methods operate per-image, retrieve from generic databases without entity indexing, and cannot ensure that an entity generated in one output is the same in the next.
In video generation, retrieval supplies motion or structure rather than appearance. RAGME~\cite{peruzzo2025ragme} injects retrieved temporal dynamics into a text-to-video model.
Animate-A-Story~\cite{he2023animate} and DreamRunner~\cite{wang2026dreamrunner} retrieve for depth maps or motion LoRA priors, while character identity is handled by per-subject optimization decoupled from the retrieval stage entirely.
Most relevant is ViMax~\cite{huang2025vimax} which retrieves references from a pool of self-generated portraits.
RAG has also been applied to video understanding~\cite{ren2026videorag, jeong2025videorag}, where retrieval optimizes for information content rather than generative conditioning.
Across all these settings, retrieval targets generic corpora, and no system retrieves appearance-preserving references from a user's own footage to condition generation.

\paragraph{Multi-shot and long-form video generation.}
Maintaining consistent characters across shots is a central challenge in long-form video generation.
Storyboard systems approach this by sharing appearance tokens across frames~\cite{dinkevich2026story2board, zhou2024storydiffusion}, propagating identity forward via causal DiT conditioning~\cite{huang2026dreamshot, wang2025echoshot}, injecting character embeddings~\cite{ma2025storynizor, zhou2026identitystory}, fine-tuning per-character LoRAs~\cite{gong2023talecrafter, wang2025autostory}, or tracking world state ~\cite{hu2024storyagent, jain2025lights, mondal2026canvas}.
These methods work well but require a manually curated set of images per character.
Multi-agent director systems~\cite{lin2023videodirectorgpt, xu2024filmagent} focus on narrative planning via LLM orchestration, but character identity is derived from text descriptions or generated portraits, not from visual references.
Memory-based approaches offer a different strategy: VideoMemory~\cite{zhou2026videomemory} and StoryMem~\cite{zhang2025storymem} maintain banks of generated keyframes, and MovieDreamer~\cite{zhao2024moviedreamer} learns autoregressive priors over visual tokens.
But this memory is built from a system's own outputs or generic video banks.
Finally, identity-preserving T2V/I2V models ~\cite{yuan2025identity, wei2024dreamvideo, jiang2024videobooth, wang2026customvideo} produce faithful single clips from provided references but generate each clip in isolation with no corpus-level memory or retrieval. 
None of these families of methods operate over a large user corpus, and none ask which frames from that corpus will most effectively condition generation.

\paragraph{AI-assisted video editing and B-roll selection.}
Editors today source B-roll through retrieval or generation, never both at once grounded in the same visual world.
Works like B-Script~\cite{huber2019b}, Transcript to Video~\cite{xiong2022transcript}, EditDuet~\cite{sandoval2025editduet}, and AutoCut~\cite{zhou2026autocut} automate B-roll assembly by retrieving and arranging existing clips from video collections into coherent timelines.
However, they cannot produce a shot that was never filmed.
Vidmento~\cite{yeh2026vidmento} takes a step further, using generative video to fill narrative gaps in a user's timeline by conditioning each new clip on its immediate neighbors (the previous and next shot) to achieve local continuity. This local conditioning suffices for bridging adjacent scenes but does not build a persistent representation of characters or settings across the project.
We target this gap and generate B-roll that is grounded in specific characters, settings, and visual style of a given source footage, with retrieval and generation operating jointly over a structured entity memory rather than over timeline neighbors alone.

\section{\method{} for Memory-Guided B-roll Generation}

\begin{figure*}[t]
    \centering
    \includegraphics[width=\textwidth]{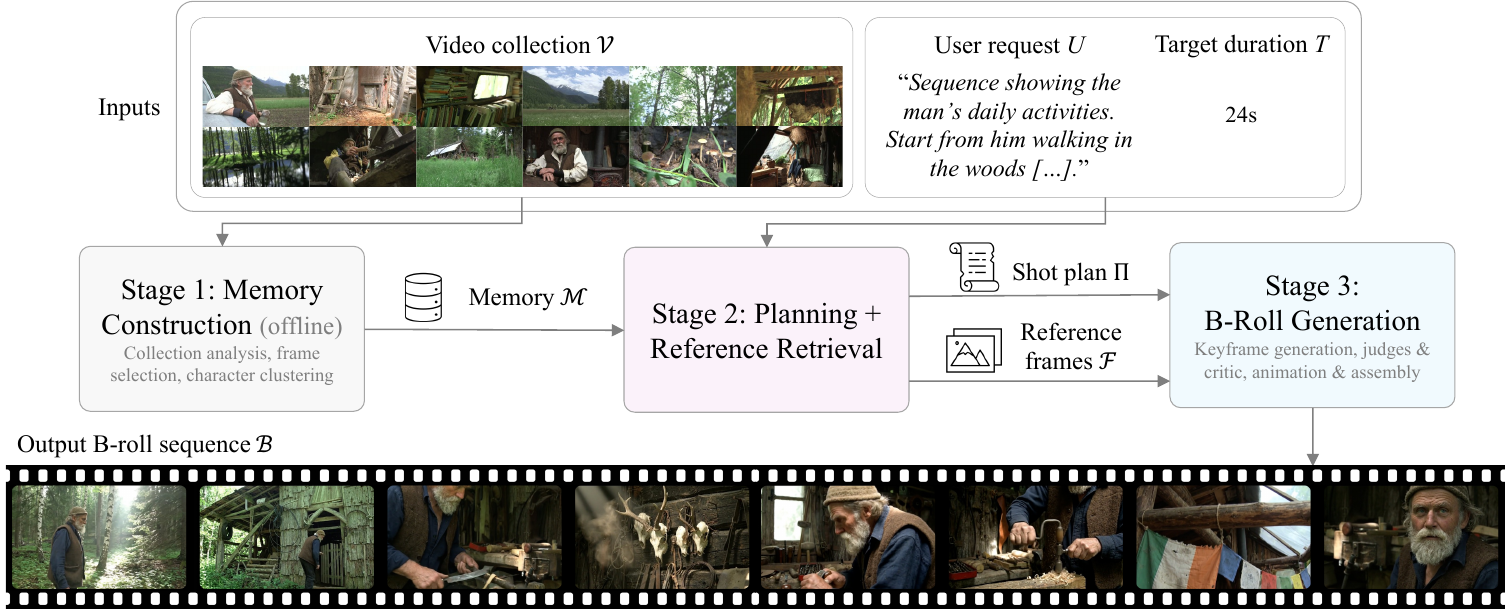}
    \caption{{\bf Overview of \method{}.} Given a captured user video collection, user request, and target duration, \method{} outputs a B-roll sequence that adheres to the inputs. \method{} comprises three stages: memory construction (performed offline and once for the video collection; Section~\ref{sec:memory}), planning and reference retrieval (Section~\ref{sec:retrieval}), and finally B-roll generation (Section~\ref{sec:generation}). See the appendix for illustrations of the different stages. Collection thumbnails are copyrighted and belong to EditStock.}
    \label{fig:overview}
\end{figure*}

Generating B-roll faithful to a user's video collection $\mathcal{V}$ entails grounding every shot in the visual world of $\mathcal{V}$: characters, settings, and style, even in configurations that do not exist in the corpus.
Given a collection $\mathcal{V}$, user request $U$, and target duration $T$, we seek to generate a sequence of B-roll $\{\mathcal{B}^1, \dots,\mathcal{B}^N\}$. 
Our key observation is that modern video generators already accept the right interface for this task: given a text prompt $p_{vid}^i$ and a set of reference frames $\mathcal{F}^i$, a video model $\mathcal{G}_{vid}$ produces a clip $\mathcal{B}^i = \mathcal{G}_{vid}(p_{vid}^i, \mathcal{F}^i)$. The challenge is then to produce the right \textit{prompt} and \textit{reference frames} for each shot; ones that will cause the generator to produce footage visually grounded in $\mathcal{V}$. This factoring drives our method, \method~ (see \Cref{fig:overview}), a system with three stages:
we (1) \textit{build a memory} $\mathcal{M}$ of $\mathcal{V}$ offline (\textsection\ref{sec:memory}), (2) use memory $\mathcal{M}$ to \textit{plan and retrieve} the reference frames $\mathcal{F}^i$ for each B-roll $\mathcal{B}^i$ (\textsection\ref{sec:retrieval}), and (3) \textit{generate the sequence} using a shot-level dual judge and a sequence-level critic that automatically assess and refine the sequence (\textsection\ref{sec:generation}).

\subsection{Memory Construction} \label{sec:memory}

We seek to build a memory $\mathcal{M}$ of the collection $\mathcal{V}$. Unlike conventional RAG systems that index \textit{moments} relevant to visual understanding tasks such as VQA, the primary goal of the memory construction stage is to distill raw, temporally redundant video into a dense, entity-centric visual reference bank comprising characters and settings usable for grounding generation. To achieve this, our approach operates in three steps. First, \textit{collection analysis} extracts structural metadata and temporal boundaries to summarize the available content. Second, \textit{frame selection} aggressively prunes the video down to a high-quality subset of reference frames that balance setting context with view-diverse character appearances. Finally, we produce global \textit{character sheets} that resolve local character detections across discrete clips into unified identities, ensuring generative consistency regardless of where the character appears in the source material. Note that memory $\mathcal{M}$ is built once per collection and reused across all subsequent requests.

\paragraph{Collection analysis.}
For each clip $v\in\mathcal{V}$, we use a vision-language model (VLM) to extract structured metadata including a clip summary, props, locations, and actions. Crucially, when extracting character information, we explicitly decouple {\em permanent traits} (e.g., core facial structure, ethnicity, build) from {\em transient traits} (e.g., clothing, accessories, momentary expressions). By embedding this permanent/transient distinction directly into the memory structure, our system is later able to enforce strict identity consistency across generated shots while still allowing for narrative-driven contextual changes (\textsection\ref{sec:generation}).
Per-clip records are aggregated into a collection-level summary $S$ describing the notable characters, settings, props, and actions in collection $\mathcal{V}$, which provides global context about what content is available in the collection.

\paragraph{Frame selection.}
A single frame rarely serves both identity and setting equally well. 
For instance, an extreme close-up carries rich facial detail but little setting context, while a wide establishing shot captures the environment but provides limited identity detail.
Frame selection must therefore balance these competing needs.
We allocate a per-clip frame budget $b^{(v)} = \lceil d^{(v)} / \rho \rceil$ to keep the memory $\mathcal{M}$ compact, where $d^{(v)}$ is the duration of clip $v$ in seconds, and $\rho$ is a target seconds per frame. We then split this budget between character- and setting-supporting frames within each clip.

We uniformly sample frames across the clips and use a VLM to score each frame for face quality and setting quality on a scale of 1-10, and assign a view-type label (e.g., frontal, profile, full-body) for any detected face. For clips containing characters (as determined by the clip-level analysis), we allocate half of the budget to face frames and half to setting frames. We filter using the face quality score, non-maximum suppression, and view-type diversification. For clips without characters, we filter based on the setting score. 
The retained frames across all clips form $\mathcal{I}$, with corresponding embeddings $\mathcal{Z}=\mathcal{E}(\mathcal{I})$ using an off-the-shelf multimodal encoder $\mathcal{E}$.
See the appendix for more details on the filtering process.

\paragraph{Character sheets.}
Per-clip metadata identifies characters within each clip but treats every clip independently, so the same person can appear under different local IDs across clips and a character query has no way to disambiguate cross-clip correspondences. We resolve identity by building a character sheet that consolidates the discovered characters. The sheet can be referenced at inference, allowing users to specify target characters. 

We run an off-the-shelf face-recognition model on every frame in the retained frames $\mathcal{I}$ to extract a face embedding, bounding box, detection score, age, and gender for each detected face. We then apply agglomerative clustering using a cosine similarity threshold on the face embeddings to group faces into candidate identities. Low-confidence clusters with fewer than 2 frames are discarded. Wrong faces can still leak into the clusters near the threshold, so we apply a verification pass where each face is compared against that cluster's highest quality face, and any faces below a similarity threshold are filtered out. The surviving clusters form the character sheet:
\begin{equation}
    C = \{(c,\mathcal{I}_c, m_c)\}_{c=1}^{|C|},
\end{equation}
where $c$ is the global character ID, $|C|$ is the number of characters, $\mathcal{I}_c \subseteq \mathcal{I}$ is the verified frame pool for that character, and $m_c$ is the per-character metadata (description, age, gender, clip count, and a mapping to local IDs assigned during analysis). 

The final memory $\mathcal{M}=(S, C, \mathcal{I}, \mathcal{Z})$ is composed of the collection summary $S$, character sheet $C$, frames $\mathcal{I}$, and corresponding image embeddings $\mathcal{Z}$.

\subsection{Planning and Reference Retrieval} \label{sec:retrieval}

The goal of the retrieval stage is to bridge the gap between the user's high-level text directive and the visual conditioning required by generative models. A text prompt alone is insufficient to guarantee that a specific character or location from the corpus will be faithfully reproduced. Therefore, our approach divides this problem into two tasks. First, {\em narrative and assembly planning} acts as an automated storyboard artist, decomposing the user's request into a structured, shot-by-shot sequence with assigned camera movements and durations. Second, {\em reference retrieval} acts as a visual researcher, querying the previously constructed memory bank to pull the exact frames needed to visually anchor each shot, prioritizing identity preservation and view diversity over semantic similarity.

\paragraph{Narrative \& assembly planning.}

We use an LLM to decompose the user request into a detailed shot plan. We separate this into a two-stage process because narrative and cinematography require different reasoning: what the story needs (which characters, setting, how long) should be decided before how to ``film'' it (actions, camera motion). Coupling them in a single pass causes the LLM to sacrifice story progression for visual appeal or vice versa.

The narrative planner receives the user request $U$, target duration $T$, and global context from memory $(S, C)$, and outputs a shot plan. Each shot description $\pi^i$ specifies which character appears, retrieval queries for those characters and the setting, a keyframe generation prompt $p_{img}^i$, and a duration $d^i$ (with $\sum d^i = T$). Retrieval queries are constrained to \textit{permanent} visual identifiers (e.g., facial structure for characters, architecture for settings) rather than \textit{transient} states (e.g., clothing, lighting). This decouples identity from state at retrieval time: frames of a character or setting can be retrieved and used as visual references for shots that depict them in different clothing or lighting than the retrieved frames showed.

The assembly planner then takes the narrative outline and assigns cinematic details, including shot type (e.g., close-up, medium, wide), camera motion (e.g., static, pan, orbit), and a video generation prompt $p_{vid}^i$, by applying guidelines derived from film editing principles (e.g., Walter Murch's ``Rule of Six''). The assembly planner may also adjust shot durations slightly for pacing. The final assembled plan $\Pi$ drives both retrieval and generation downstream.

\paragraph{Reference retrieval.}
For each shot $i$, we retrieve a per-shot reference set $\mathcal{F}^i$ from memory $\mathcal{M}$ using the shot details $\pi^i$. Each query $q$, whether character or setting, is embedded using the same multimodal encoder $\mathcal{E}$ from \textsection\ref{sec:memory} and ranked against candidate frames via cosine similarity $\mathrm{sim}(\mathcal{E}(q), z)$ for $z \in \mathcal{Z}$. However, standard top-$k$ ranking has three main limitations in this setting: (i) character queries must resolve to the correct physical person, but textual descriptions of a character (even ones based on permanent traits) can match other characters who share similar traits, (ii) view type matters for face identity in a way that raw similarity does not capture (frontal views carry the strongest identity signal but often lose similarity rankings to wider views), and (iii) the cause of weak retrievals can be ambiguous--either the collection genuinely lacks the requested context, or the query phrasing failed to retrieve it.

We address these limitations with additional constraints on the retrieval process. First, character-gated retrieval restricts each character query  to the verified frame pool $\mathcal{I}_c$ for character $c$. Specifically, the candidate set for $q_{char}^{i, c}$ is $\{\mathcal{E}(f) : f \in \mathcal{I}_c\}$ rather than $\mathcal{Z}$, so similarity over text-image embeddings cannot route the query to a different person. Second, per-character queries $q_{char}^{i, c}$ are ranked and selected separately from the top-$k$ candidates above a similarity floor, allowing up to two frames per character under a view-diversity rule that prioritizes a frontal view when available. Setting queries are ranked over all of $Z$, selecting up to two frames per shot. When a query does not return any matches above a similarity threshold, the collection summary $S$ disambiguates whether the requested context exists in $\mathcal{V}$; if it does, the query is rewritten and retried. The final set of frames forms the per-shot reference $\mathcal{F}^i$.

\subsection{B-Roll Generation} \label{sec:generation}

The goal of the generation stage is to synthesize the final video sequence while adhering to the assembled shot plan $\Pi$ and preserving the visual identity of the source collection $\mathcal{V}$ through the retrieved reference sets $\{\mathcal{F}^i\}_{i=1}^N$. Since high-fidelity video generation is computationally expensive, we decouple the generation process into two stages. First, we synthesize a sequence of text-anchored representative keyframes to ensure shot-to-shot continuity. We then evaluate and refine these keyframes through an automated multi-scale feedback loop that checks both local shot quality and global sequence coherence. Only after the keyframe sequence satisfies these constraints do we render the final video clips.

\paragraph{Context-aware keyframe generation.}
Reference frames retrieved for a single shot typically come from clips in which the character appears under different clothing, accessories, and lighting. Additionally, unlike curated single-subject images that prior reference-based generation methods typically use for grounding, real corpus frames often carry extraneous people, props, and background detail. Passing these frames directly to $\mathcal{G}_{img}$ with prompt $p_{img}^i$ requires the model to infer which attributes from each reference frame should transfer to the generated keyframe. 

We address this ambiguity with text ``anchors'' that describe the character and setting state entering each shot, distinguishing permanent traits (e.g., face structure) from transient ones (e.g., clothing).
We augment the original prompt with the anchors and planned shot type to form $\tilde{p}_{img}^i$, which directs $\mathcal{G}_{img}$ to carry permanent attributes across shots while allowing transient attributes to be conditioned by the current shot's references and prompt. Additionally, to preserve continuity across shots, each generated keyframe enters the next shot as an additional reference. The keyframe for shot $i$ is then:
\begin{equation}
    \tilde{x}^i = \mathcal{G}_{img}(\tilde{p}_{img}^i, \mathcal{F}^i, \tilde{x}^{i-1}),
\end{equation}
with $\tilde{x}^0 = \varnothing$ for the first shot. After each successful keyframe, transient anchors are updated to reflect any state evolution observed in the new keyframe. Any conflicts between the updated anchor and existing state (e.g., a character's shirt changes mid-sequence) are recorded and surfaced to the sequence critic below.

\paragraph{Local shot verification.}
We define two judges to evaluate each shot's visual alignment with $\mathcal{V}$: a hybrid character judge and a setting judge. 
For each character $c$ in keyframe $i$, the character judge $\mathcal{J}_{char}$ produces two scores: a face-centric score $s_{F, c}^i$ using cosine similarity between the embeddings of the detected face in $\tilde{x}^i$ and the highest quality frontal face from the retrieved character pool, and an appearance score $s_{A, c}^i \in [1, 10]$ from a VLM evaluating attribute consistency. These are combined as:
\begin{equation}
    s_I^i = \frac{1}{|l_{char}^i|} \sum_{c \in l_{char}^i} [w_c \cdot 10 s_{F, c}^i + (1 - w_c) \cdot s_{A, c}^i],
\end{equation}
where $w_c$ is a weight that depends on the face area relative to the frame and the yaw angle, and $l_{char}^i$ is a list of character IDs. $w_c$ is higher in close-ups where the face dominates the frame, lower in wide shots where the face is small and face embeddings become unreliable, and 0 for extreme viewpoints (e.g., near-profile) where face embeddings are unstable. The character judge is skipped if no characters appear in the keyframe.

The setting judge $\mathcal{J}_{set}$ is a VLM that produces a score $s_{S}^i \in [1, 10]$ for how well the generated setting adheres to the retrieved setting references. The two judges combine into a per-keyframe quality:
\begin{equation}
    s_{quality}^i = w_{type} \cdot s_I^i + (1 - w_{type}) \cdot s_S^i,
\end{equation}
where $w_{type}$ is a weight factor dependent on shot type, following similar logic as $w_c$ where characters dominate tight shots and setting dominates in wide shots. If the weighted combination falls below a given threshold, the keyframe is regenerated with an updated prompt using judge feedback up to a fixed number of times before proceeding onto the next keyframe.

\paragraph{Global sequence-level critic.}
Per-shot judges catch local failures, but several failure modes (e.g., repetitive shot grammar, character/setting drift between shots) only emerge when the sequence is viewed as a whole and cannot be detected by any single shot's score. After all keyframes in the planned sequence are generated, a sequence critic $\mathcal{C}$ examines the keyframe sequence $\{\tilde{x}^i\}_{i=1}^N$ together with the user request $U$, target duration $T$, plan $\Pi$, and accumulated state conflicts surfaced during anchor updates. The critic labels each shot with one of $\{\texttt{keep},\allowbreak \texttt{regenerate},\allowbreak \texttt{remove},\allowbreak \texttt{retime},\allowbreak \texttt{add}\}$, 
allowing the sequence to be edited as needed. The critic loop runs until every action is \texttt{keep} or until a max number of iterations are reached.

\paragraph{Keyframe animation and assembly.}
Once the critic either accepts the keyframe sequence or the maximum number of iterations is reached, each keyframe $\tilde{x}^i$ is animated to a clip $\mathcal{B}^i = \mathcal{G}_{vid}(\tilde{p}_{vid}^i, \tilde{x}^i)$ using the planned video generation prompt $\tilde{p}_{vid}^i$ augmented with the camera motion. We route between two conditioning modes per shot based on the camera motion assigned by the assembly planner: first-frame image-to-video (I2V) is used for camera motions like static or zoom, where the keyframe contains most or all of the view the clip will reveal, whereas reference-based I2V is used for camera motions like orbit or tracking, where the movement exposes off-screen area. For the latter, we add the highest-scoring character and setting reference frames from $\mathcal{F}^i$ as additional image conditioning. Each clip $\mathcal{B}^i$ is trimmed to its planned duration $d^i$ by removing the tail and then concatenated to produce the final B-roll sequence $\{\mathcal{B}^i\}_{i=1}^N$.

\section{Experiments} \label{sec:experiments}
We implement \method~using Gemini 3 Flash for memory construction and narrative planning, Gemini 3.1 Pro for visual judges and critics, and Veo 3.1 for video generation. We provide implementation details: specific model variants, embedding architectures, hyperparameters, and prompts, in the appendix.

\subsection{Baselines}
As discussed in Section~\ref{sec:related_work}, no prior method addresses corpus-grounded B-roll generation. We thus set up two baselines:

\paragraph{Text-to-Video Generation with Planning (T2V + Planner).} 
This baseline generates the sequence from the user request U and target duration T, without access to the user's video collection $\mathcal{V}$. Since Veo 3.1 generates short clips, we use a single-stage planner, implemented with Gemini 3 Flash, to decompose U and T into a sequence of shot descriptions before generation. Each planned shot is generated with Veo 3.1. Unlike our method, this baseline does not split planning into narrative and assembly stages and does not use retrieved references or intermediate keyframes. It therefore serves as a baseline for what state-of-the-art text planning and video generation models can achieve without access to the user’s video collection.

\paragraph{EditDuet~\cite{sandoval2025editduet}} An agentic editing framework that retrieves and assembles real clips from a user's video collection based on a text request. Because it lacks a video generation component, comparing against EditDuet isolates the value of our generative pipeline in fulfilling directives where the required footage was either unable to be retrieved or never explicitly captured.

\subsection{Evaluation Dataset and Metrics}
\paragraph{Dataset.} We use the evaluation dataset from EditDuet \cite{sandoval2025editduet}, consisting of 5 documentary projects with raw footage from EditStock \cite{editstock} and manually annotated high-level user requests describing the B-roll sequences in the final cuts from each project. We randomly sample up to 9 user requests per project (39 in total), spanning diverse styles (e.g., serene, nature-oriented vs. fast-paced, action-filled), with an average target duration of 19.1s. EditDuet also optionally accepts audio transcript as input, which we provide to EditDuet where available in the dataset.

\paragraph{Metrics.}
We evaluate performance across three dimensions:\\
\noindent \textit{(1) Duration Control:} We report the absolute difference $|\Delta T|$ (in seconds) between the output sequence duration and target duration $T$.\\
\noindent \textit{(2) Prompt Adherence:} We conduct a forced-choice A/B user study to evaluate narrative fidelity. We display the input text request $U$ alongside a pair of generated B-roll sequences (ours and a baseline, in randomized order) and ask evaluators: ``Which video better represents the text prompt above?"\\
\noindent \textit{(3) Visual Alignment:} We conduct a second A/B user study to evaluate corpus fidelity. To familiarize the evaluators with the visual world of the source footage, we display a static reference board of 5–6 images taken from the project's EditStock overview page. We then ask: ``Which video better fits in with the reference frame collection shown above?" 

For the A/B studies we recruited 286 participants via Prolific, who provided on average 5 unique ratings per video pair. We opted for A/B testing, a common protocol for generation, over Likert scale evaluations due to the challenges of calibrating scores across participants. We report \textit{win rate}, the percentage of users preferring \method~ over the alternative approach.

\subsection{Results}

Table~\ref{tab:baselines} summarizes our quantitative results. \method~achieves better duration control and prompt adherence while remaining visually aligned with the source footage.

\paragraph{Duration Control.} \method~achieves near-perfect adherence to the target duration $T$, with an average absolute error of just $0.12 \pm 0.05$ seconds ($0.8\%$ relative error). In contrast, EditDuet struggles with precise timing ($4.77 \pm 4.75$ seconds error) because it is bottlenecked by the lengths of the available retrieved raw clips. The T2V + Planner baseline also exhibits higher variance ($1.26 \pm 0.95$ seconds error), largely because failed video generations are excluded from the final sequence. In contrast, \method~can fall back to a still keyframe for the planned shot duration, enabling more reliable duration control even when generation fails.

\paragraph{Prompt Adherence.} \method~overwhelmingly outperforms both baselines in fulfilling the user's text directive. Against EditDuet, our full system achieves a 94.5\% win rate. Because EditDuet can only assemble existing footage, it fundamentally fails when a user requests a specific action or perspective that was never captured. \method’s generative pipeline gracefully overcomes this limitation. Furthermore, \method~achieves a 60\% win rate over the T2V + Planner baseline. Although both methods synthesize new content, \method's keyframe chaining and iterative feedback ensures that the multi-shot sequence accurately reflects the prompt across temporal cuts without drifting.

\paragraph{Visual Alignment.} The primary goal of our system is to synthesize footage that reflects the user's specific visual world. When compared to T2V + Planner, which generates purely from text without corpus grounding, human evaluators preferred \method~92.8\% of the time, validating the need for our memory and retrieval architecture.
EditDuet is a stricter baseline: it constructs sequences directly using footage in the user's collection, which we would expect to have a natural advantage in visual alignment. Even so, evaluators preferred \method's generated sequences in 41.8\% of comparisons, indicating that generation quality together with collection-grounded memory and retrieval can narrow the perceived alignment gap between generated B-roll and real captured footage.

\begin{table}[]
    \caption{Comparison against baselines from A/B user studies.
    The green bar and  number indicates the win rate for our approach, and the red bar indicates the win rate for the baseline. * indicates a p-value $<0.02$.}
    \resizebox{\linewidth}{!}{%
    \begin{tabular}{cccc}
        \toprule
        Method & $|\Delta T|$ (s) $\downarrow$ & Prompt adherence ($U$) &  Visual alignment ($\mathcal{V}$)\\
        \midrule
        T2V + Planner & 1.26 $\pm$ 0.95 & \prefbar{40.0}* & \prefbar{7.2}* \\
        EditDuet & 4.77 $\pm$ 4.75 & \prefbar{5.5}* & \prefbar{58.2}\phantom{*} \\
        \method & \textbf{0.12 $\pm$ 0.05} & N/A & N/A \\
        \bottomrule
    \end{tabular}
    }
    \label{tab:baselines}
\end{table}

\subsection{Ablation Study}

We ablate the three stages of \method~to isolate their contributions. Each variant modifies one stage and leaves the rest in its default configuration. Table~\ref{tab:ablation} reveals several key insights:

\paragraph{Memory construction requires quality-filtering, not just temporal sampling.} Replacing our frame selection with naive sampling strategies degrades visual alignment. When compared to \textit{Uniform} sampling or \textit{Scene-detected} boundary frames, human evaluators preferred \method's full memory for visual alignment (78.2\% and 83.3\% win rates, respectively), confirming our hypothesis that generative models require specific, high-quality conditioning frames of entities rather than just semantically relevant moments. Also, using unfiltered uniformly sampled frames to build the character sheet (\textit{Uniform + }$C$) also led to a drop in visual alignment.

\paragraph{Retrieval must be entity-gated, not just semantically matched.}  Replacing our retrieval module with
\textit{Caption-RAG} (text-text similarity search using clip descriptions and taking the mid-frame from matching clips) often retrieves frames that semantically match the prompt's action but feature the wrong entity, resulting in a sharp drop in both prompt adherence (63.4\% win rate for \method) and visual alignment (54.3\%). Removing the character ID gate (\textit{No char. ID gate}) also results in a drop in both Prompt adherence and Visual alignment, showing that strict entity-gating is necessary to prevent the generator from mixing identities.

\paragraph{Iterative feedback is essential for sequence-level coherence.} 
Disabling the per-shot judges (\textit{w/o judges}) reduced the system's ability to maintain strict visual alignment (59.7\% win rate for \method), as the generator was allowed to pass localized identity failures into the final assembly. Disabling the sequence-level critic (\textit{w/o critic}) hurt prompt adherence (65\% win rate for \method), demonstrating that while individual shots might be locally coherent, a global critic is required to catch narrative stalls, missed actions, or pacing issues across the compiled sequence.




\begin{table}[]
    \caption{Ablation study results from A/B user studies. * indicates a p-value $<0.02$.}
    \resizebox{\linewidth}{!}{%
    \begin{tabular}{llcc}
        \toprule
        Stage & Variant & Prompt adherence ($U$) & Visual alignment ($\mathcal{V}$) \\
        \midrule
        \multirow{3}{*}{Memory}
          & Uniform & \prefbar{53.8}\phantom{*} & \prefbar{21.8}* \\
          & Scene-detected & \prefbar{47.3}\phantom{*} & \prefbar{16.7}* \\
          & Uniform + $C$ & \prefbar{38.6}* & \prefbar{32.2}* \\
        \midrule
        \multirow{3}{*}{Retrieval}
          & Random & \prefbar{28.8}* & \prefbar{46.0}\phantom{*} \\
          & Caption-RAG & \prefbar{36.6}* & \prefbar{45.7}\phantom{*} \\
          & No char. ID gate & \prefbar{40.2}* & \prefbar{44.2}\phantom{*} \\
        \midrule
        \multirow{2}{*}{Generation}
            & w/o judges & \prefbar{52.1}\phantom{*} & \prefbar{40.3}* \\
            & w/o critic & \prefbar{35.0}* & \prefbar{43.6}\phantom{*} \\
        \bottomrule
    \end{tabular}
    }
    \label{tab:ablation}
\end{table}

\subsection{Qualitative Results} 
\begin{figure*}[t]
    \centering
    
    \begin{subfigure}[t]{\linewidth}
    \includegraphics[width=\linewidth]{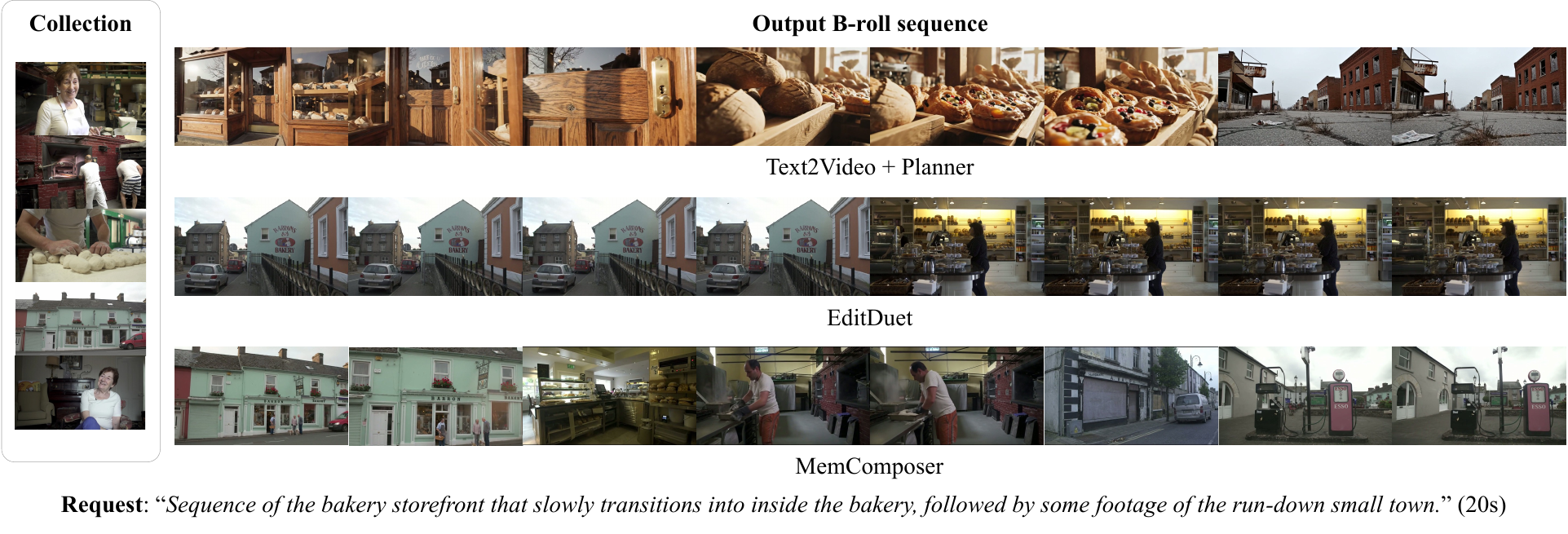}
    \label{fig:comp01}
    \end{subfigure}
    \par\vspace{-1.0em}
    \noindent\makebox[\linewidth]{\rule{\linewidth}{0.4pt}}
    \par\vspace{0.5em}
    \begin{subfigure}[t]{\linewidth}
    \includegraphics[width=\linewidth]{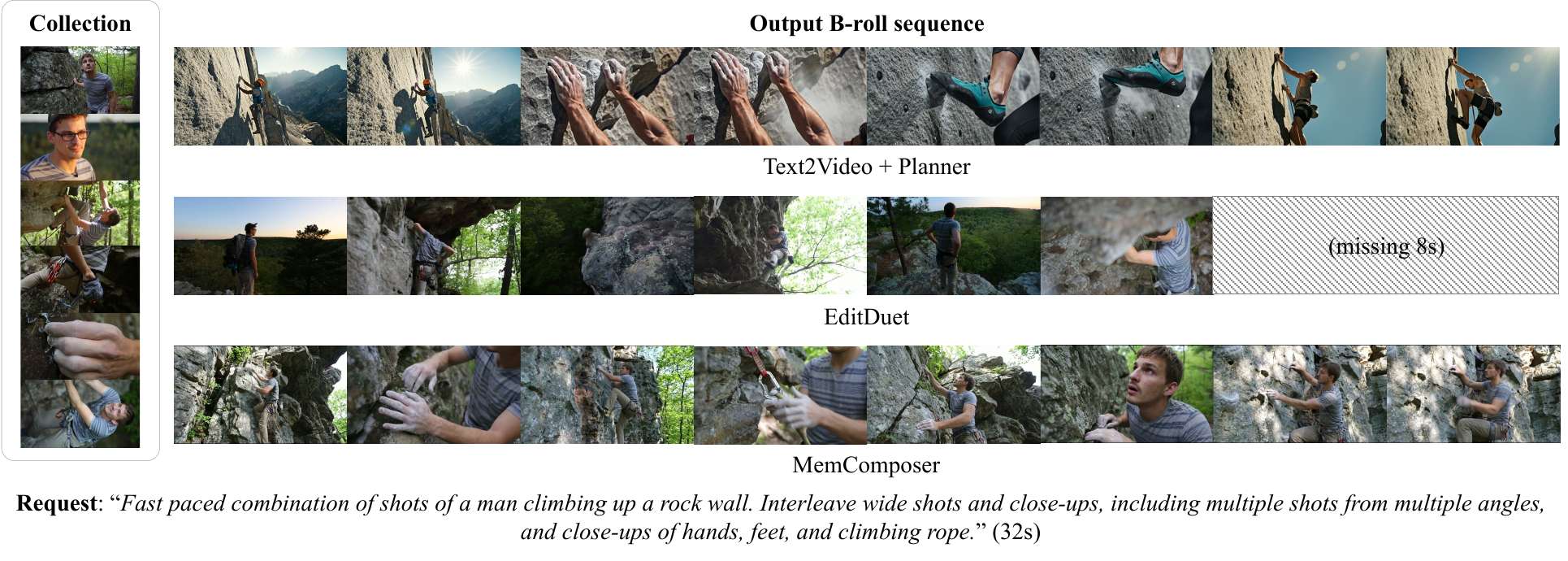}
    \label{fig:comp02}
    \end{subfigure}
    \vspace{-2.5em}
    \caption{{\bf Qualitative comparison.} We compare \method{} to the baselines on two collections. \method{} better matches the setting from the input collection than Text2Video+Planner and better adheres to the user request than EditDuet. Collection thumbnails are copyrighted and belong to EditStock.}
    \label{fig:comparison_01}
\end{figure*}

\begin{figure*}[t]
    \centering
    \includegraphics[width=\textwidth]{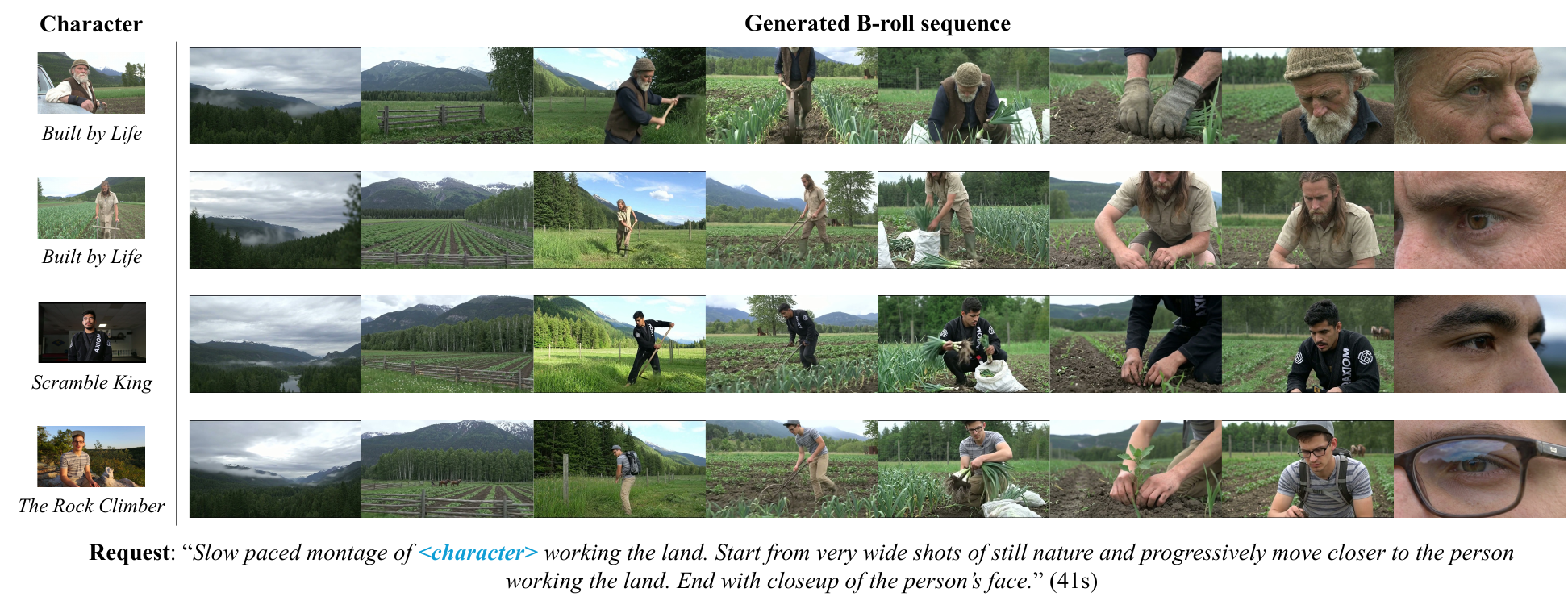}
    \vspace{-2.5em}
    \caption{{\bf Qualitative result of swapping characters.} For the {\it Built by Life} collection, we swap in different characters (spanning different projects). Notice how each row maintains the setting from {\it Built by Life} while correctly depicting the target character. Collection thumbnails are copyrighted and belong to EditStock.}
    \label{fig:character-swap}
\end{figure*}

\begin{figure*}[t]
    \centering
    
    \begin{subfigure}[t]{0.48\linewidth}
    \includegraphics[width=\linewidth]{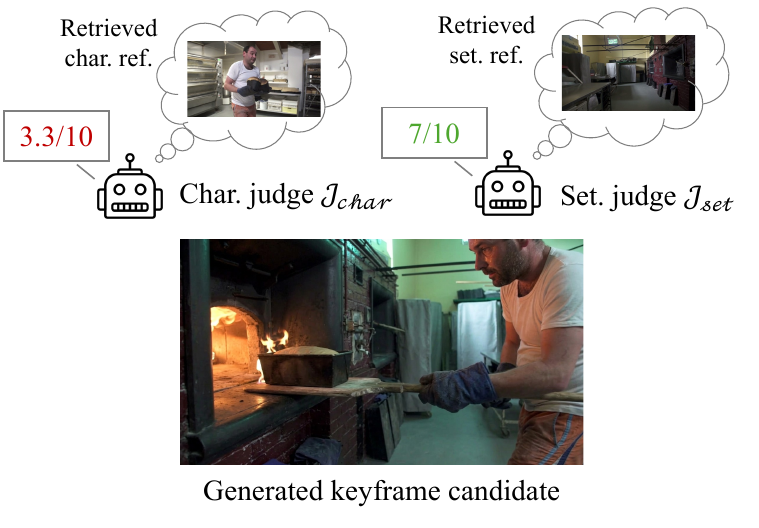}
    \caption{Judges outputs.}
    \label{fig:judges}
    \end{subfigure}
    \hfill
    \begin{subfigure}[t]{0.48\linewidth}
    \includegraphics[width=\linewidth]{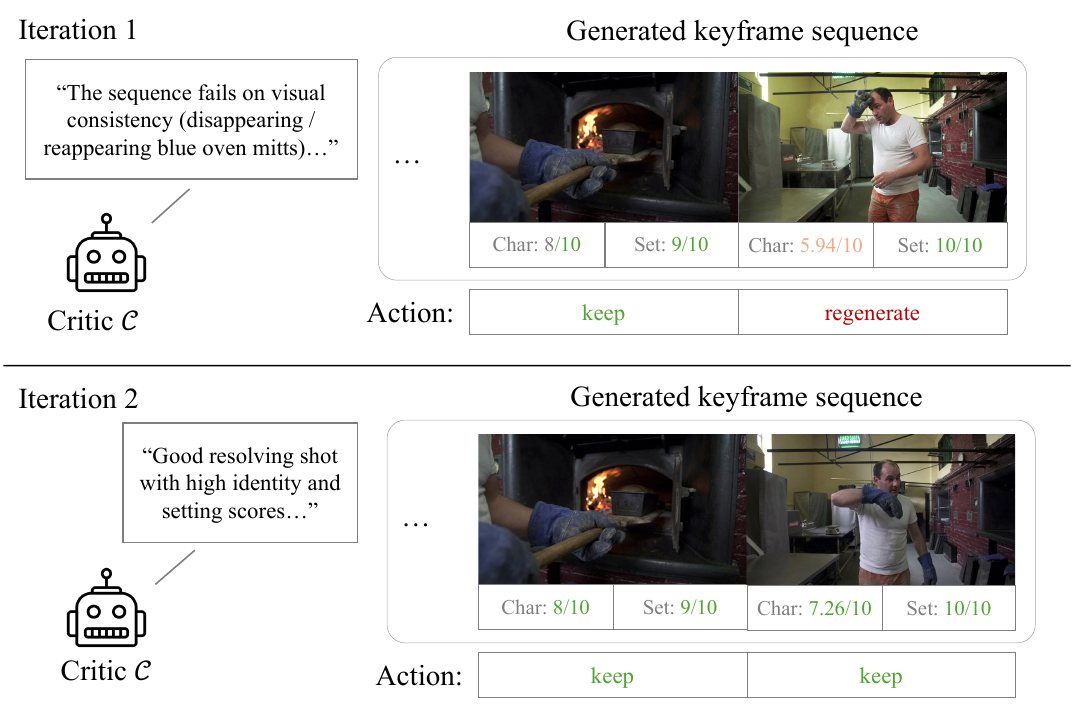}
    \caption{Critic outputs.}
    \label{fig:critic}
    \end{subfigure}

    \caption{{\bf Judges and critic outputs.} (a) We illustrate the outputs from judges $\mathcal{J}_{char}$ and $\mathcal{J}_{set}$ over the generated keyframe candidate, with scores out of 10. While the generated keyframe passes the setting judge $\mathcal{J}_{set}$, it does not pass the character judge $\mathcal{J}_{char}$. (b) We illustrate the critic $\mathcal{C}$ over two generation iterations. In the first iteration, the critic deems that the second keyframe fails due to visual consistency. In the second iteration, both keyframes are deemed acceptable by the critic. The retrieved reference images are copyrighted and belong to EditStock.}
    \label{fig:judges-critic}
\end{figure*}

\begin{figure*}[t]
    \centering
    \includegraphics[width=0.9\textwidth]{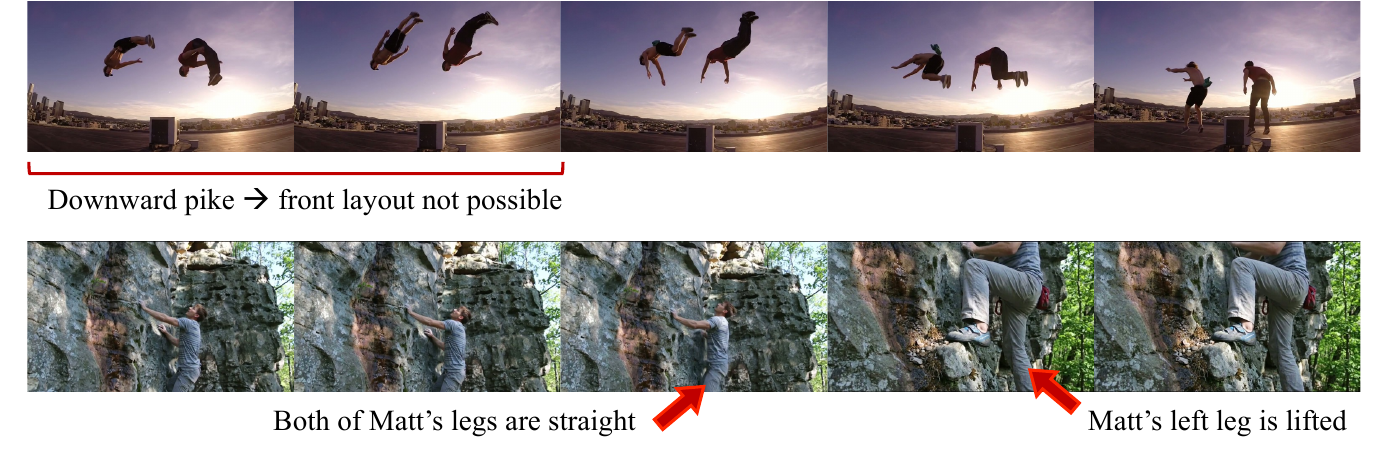}
    \caption{{\bf Example failures.} Example failures include (top) implausible motion due to complex physics and (bottom) fine-grained scene characteristics are not enforced to be consistent across shots.}
    \label{fig:failures}
\end{figure*}

Please watch the companion video to properly view the results. 
Figures~\ref{fig:comparison_01}-\ref{fig:failures} are static snapshots and are best judged via the video.

\paragraph{Generating missing footage.} Figure~\ref{fig:comparison_01} shows \method~ generating shots that were not captured. 
In the bakery example (top), \method~ starts the sequence with storefront shots then cuts to interior shots of the bakery, and ends with the town; all shots retain the look of the original collection. 
The Text-to-Video+Planner baseline produces plausible but generic footage. EditDuet stays grounded, because it uses real clips, but it is limited to what it can retrieve.
In the climbing example (bottom), \method~produces a complete montage with varied shot scales, while EditDuet cannot cover the full requested sequence. 
Overall, \method~combines the strengths of both baselines: it can synthesize missing coverage and preserve the visual identity of the source collection.

\paragraph{Swapping characters while preserving the setting.}
Figure \ref{fig:character-swap} shows the same rural (\textit{Built by Life}) setting reused with different target characters. 
The landscape, fields, and overall visual style remain the same, while the subject changes.
We enable this application by injecting the target character sheet into the \textit{Built by Life} memory: when a character ID is specified, retrieval is gated to the pool of frames labeled as containing that character.
This allows \method~to preserve the grounded setting while independently controlling which character appears in it.

\paragraph{Catching failures during generation.}
Figure \ref{fig:judges-critic} shows a representative judge-and-critic loop in action.
A keyframe can satisfy the setting judge yet fail identity checks.
The critic then catches cross-shot issues, such as blue oven mitts disappearing/reappearing, and accepts the sequence after regeneration. 
This qualitative result illustrates why \method~ needs both local judges and sequence-level critic to achieve identity fidelity and sequence consistency.

\paragraph{Failure modes.}
Figure \ref{fig:failures} illustrates example failures. The top example fails on complex physical motion.
The bottom example shows fine-grained pose inconsistency, where the climber’s leg configuration changes across shots.
These cases show that \method~improves grounding and sequence structure, but motion realism and precise video-level continuity remain open challenges.

\section{Conclusion}
We introduced \method, a system for generating collection-grounded B-roll. \method~ overcomes the scalability limits of manual curation by constructing an offline, entity-centric memory from raw footage that explicitly decouples permanent from transient traits. This enables generation-informed retrieval and a multi-scale iterative generation loop. By ensuring visual anchors are optimal for video generation conditioning, \method~consistently produces cohesive sequences that preserve the identity of the user’s visual world. Our user study shows that our method outperforms text-to-video and retrieval-only baselines.






\bibliographystyle{ACM-Reference-Format}
\bibliography{references}

\newpage

\clearpage

\appendix

\section{Limitations}

While \method~demonstrates strong performance, it is subject to the limitations illustrated in Figure~\ref{fig:failures}. First, because our multi-scale critic operates exclusively on generated keyframes to maintain computational feasibility, the system lacks a video-level feedback loop. Consequently, while individual shots are visually aligned, the cuts between clips can occasionally feel disjointed since motion patterns and compositions are not explicitly forced to bridge smoothly across boundaries. Incorporating video-level judges could address this, though at a significantly higher computational cost. However, controlling the overall pacing/dynamics of the sequence is an open question on its own.

Second, our pipeline's final fidelity is inherently bound by the capabilities of the underlying video generation model. We observe that current models occasionally struggle with complex physics and rapid actions, leading to implausible motions or localized hallucinations. Additionally, we trim clips during assembly because Veo is constrained to only producing 4/6/8 second outputs, depending on the conditioning mode.

Third, the current decomposition of characters/settings may not cover the full granularity of controllable assets. We evaluated a few examples using specific objects/props within the collections and found that the quality of the results depends on how well the memory construction stage captured the object. For example, more prominent/frequently observed objects were often saved, but smaller/infrequently seen objects were sometimes missed, highlighting the need for more explicit filtering during the memory stage.

\section{Ethical Considerations}
\label{sec:ethical_considerations}
Because \method~ utilizes face recognition and video generation to synthesize new actions for real individuals, it carries the risk of being misused to create non-consensual deepfakes or misrepresent individuals. Our system is explicitly designed as a creative editing tool for users operating over their own personal or licensed media collections. To mitigate misuse, such frameworks should be paired with robust origin-tracing protocols, such as invisible watermarking, and should be deployed alongside standard commercial safeguards that filter malicious generation prompts.

\section{Memory Construction: Frame Filtering Details}

We first uniformly sample frames across the clips and use a VLM to score each frame $f$ for face quality $\phi_{face}^{(f)}$ and setting quality $\phi_{set}^{(f)}$ on a scale of 1-10, and assign a coarse view-type label (e.g., frontal, profile, full-body) for any detected face. For clips containing characters (as determined by the clip-level analysis), half of budget $b^{(v)}$ is allocated to face frames and half to setting frames. The face half $I_{face}^{(v)}$ is built by filtering face quality $\phi_{face}^{(f)}$ below a threshold, applying non-maximum suppression in $\Delta t$-second windows to break temporal redundancy, and then preferring view-type diversity, selecting the best frame per view type before filling the remaining $\lceil b^{(v)}/2\rceil$ slots by descending face quality $\phi_{face}^{(f)}$. View diversity is enforced because frontal and three-quarter views carry complementary identity signal and the highest-similarity frames within a clip are typically the same view. The setting half $I_{set}^{(v)}$ uses the analogous procedure with its respective threshold. If $|I_{face}^{(v)} \cup I_{set}^{(v)}| < b^{(v)}$ due to overlaps between the two sets, we backfill with the remaining frames sorted by $\max(\phi_{face}^{(f)}, \phi_{set}^{(f)})$. Clips without characters allocate the full budget to setting frames.

While we do not have per-frame ground truth labels for the evaluation projects to measure character clustering accuracy, we manually inspected all character references used to generate the evaluation set output sequences and did not observe any incorrect retrievals since our retrieval prioritizes high-quality frames where characters can be reliably identified by state-of-the-art face embedding models.

\section{Implementation Details}
\label{sec:implementation_details}

\begin{figure}[th]
    \centering
    \includegraphics[width=\linewidth]{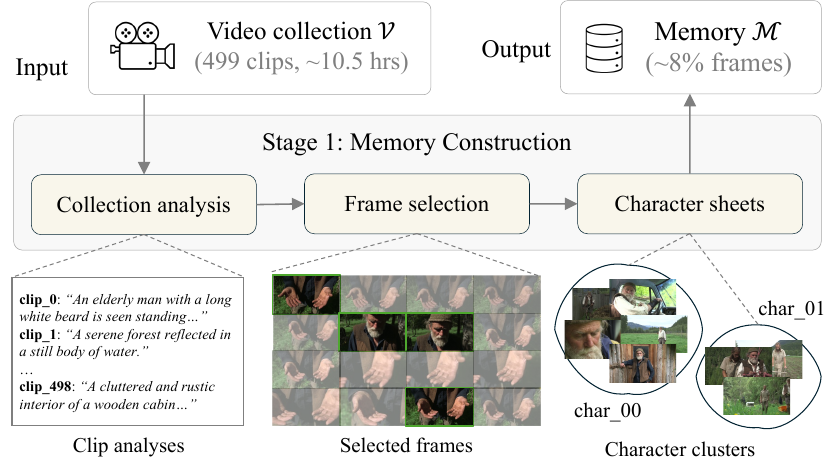}
    \caption{{\bf Stage 1: Memory construction.} The memory construction stage comprises steps for collection analysis, frame selection, and computing character sheets. The bottom row illustrates example outputs of each of the steps for this stage. All thumbnails are copyrighted and belong to EditStock.}
    \label{fig:memory}
\end{figure}

\begin{figure*}[th]
    \centering
    \includegraphics[width=0.8\linewidth]{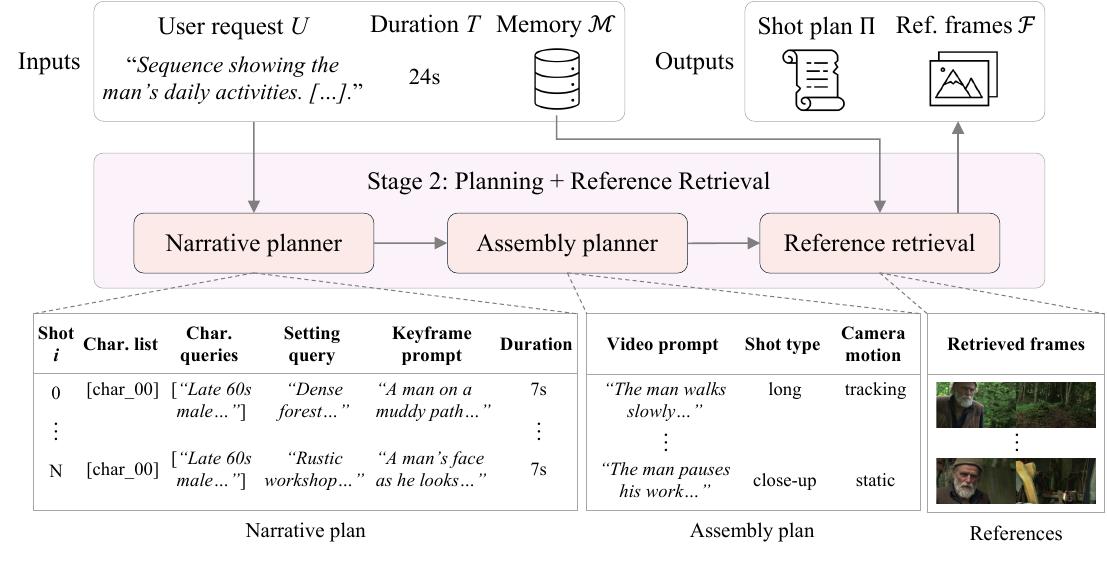}
    \caption{{\bf Stage 2: Planning and reference retrieval.} The planning and reference retrieval stage comprises steps for narrative planning, assembly planning, and reference retrieval. The bottom row illustrates example outputs of each of the steps for this stage. All thumbnails are copyrighted and belong to EditStock.}
    \label{fig:retrieval}
\end{figure*}

\begin{figure*}[th]
    \centering
    \includegraphics[width=\linewidth]{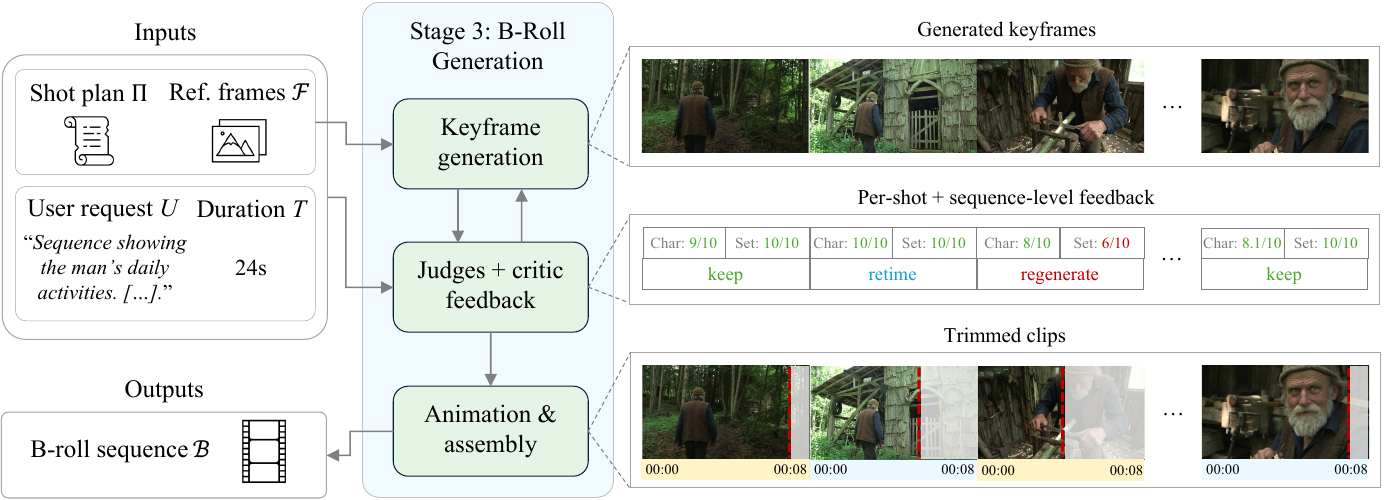}
    \caption{{\bf Stage 3: B-roll generation.} The B-roll generation stage comprises steps for keyframe generation and animation \& assembly. A set of judges and a critic gives feedback to the keyframe generation step for regenerating keyframes for improved adherence to the inputs. The right side illustrates example outputs of each of the steps for this stage.}
\label{fig:generation}
\end{figure*}

\subsection{Model Architectures}
The narrative and assembly planners ($\mathcal{P}_{narr}, \mathcal{P}_{asmb}$), collection analysis, frame annotation, and text-anchor generation all use Gemini 3 Flash. All visual judges and the sequence critic ($\mathcal{J}_{char}, \mathcal{J}_{set}, \mathcal{C}$) use Gemini 3.1 Pro. The face component of $\mathcal{J}_{char}$ uses InsightFace with ArcFace embeddings \cite{deng2018arcface}, which we also use for character clustering and character-gated retrieval. Memory frames $I$ and retrieval queries are encoded using Gemini Embedding 2. We use Gemini 3.1 Flash Image (Nano Banana 2) for $\mathcal{G}_{img}$ and Veo 3.1 for $\mathcal{G}_{vid}$. 

MemComposer takes approximately 1 min + 3 min $^\ast N$ to generate a full output sequence, where $N$ is the number of shots.

\subsection{Baseline Implementations}

\paragraph{Text-to-Video Generation with Planning (T2V + Planner).} To create a fair comparison for the T2V baseline, we implement a single-stage planner derived from our proposed two-stage planner using Gemini 3 Flash (prompt provided in \Cref{fig:t2v-planner-prompt}). We bypass the keyframe generation stage entirely and directly input the generated video prompts to Veo 3.1.

\begin{figure*}
\begin{lstlisting}[basicstyle=\small\ttfamily, breaklines=true, frame=single]
You decompose a user's b-roll request into a sequence of distinct shot
descriptions for a text-to-video model. Each shot is rendered independently
with no shared references, so every prompt must be fully self-contained:
re-state the subject, setting, and action in each one -- no pronouns referring
to prior shots, no 'same as before.' Vary what the viewer sees across the
sequence (different angles, different moments, different framings) so the
concatenated result has visual variety. Each prompt yields one continuous
take, so do not include cuts, transitions, fades, or multiple actions within
a single shot description.

B-roll is supplementary footage: no talking heads, no direct address to
camera, no interviews. People doing or interacting is fine. People looking
into the lens and speaking is not.

User request: {user_prompt}

Decompose into exactly {N} shots with these durations (seconds): {seg_durations}.
Return a JSON object with key 'shots', a list of {N} objects each with a
'prompt' field.
\end{lstlisting}
\caption{VLM prompt for the Text2Video planner.}
\label{fig:t2v-planner-prompt}
\end{figure*}

\paragraph{EditDuet.} We follow the original implementation as described by~\cite{sandoval2025editduet}, utilizing the same collection indexing and assembly protocols.

\paragraph{Alternative retrieval approaches.} We experimented with alternative retrieval strategies, such as text-text similarity (\textit{Caption-RAG} ablation) but found that our proposed text-image similarity performed better. We did not try any learning-based retrieval approaches due to limited training data.

\paragraph{Handling generation filters.} During final video rendering, Veo 3.1 may occasionally fail to produce an output clip due to its built-in safety filters. When this occurs in our \method~pipeline, we gracefully degrade by inserting a static video clip of the successfully generated keyframe for the planned duration of that shot. Because the T2V baseline operates without intermediate keyframes, any clips blocked by filters in the baseline are simply omitted from its final assembled sequence.

\subsection{Camera Shot and Motion Types}

The assembly planner can select camera shot and motion types from the following lists:
\begin{itemize}
    \item \textit{Shot types} (13): extreme close up, close up, medium close up, medium shot, medium long shot, long shot, extreme long shot, over-the-shoulder, point-of-view, aerial, low-angle, high-angle, dutch-angle
    \item \textit{Camera motions} (14): static, pan left/right, tilt up/down, dolly in/out, tracking, crane up/down, orbit, handheld, zoom in/out
\end{itemize}

\subsection{Hyperparameters}

The hyperparameters used the the various stages of our pipeline can be found in \Cref{tab:hyperparameters}.

\begin{table*}
    \caption{Hyperparameters used in our pipeline.}
    \label{tab:hyperparameters}
    \small
    \begin{tabular}{l l c p{7.5cm}}
        \toprule
        \textbf{Stage} & \textbf{Hyperparameter} & \textbf{Value} & \textbf{Description} \\
        \midrule
        \multirow{12}{*}{Memory}
            & \texttt{SAMPLE\_INTERVAL} & 2 & Base frame sampling interval (sec) \\
            & $\rho$ & 15 & Target seconds-per-frame for frame budgeting \\
            & \texttt{MIN\_CANDIDATES} & 6 & Min.\ number of candidate frames to score before filtering \\
            & \texttt{MIN\_FACE\_QUALITY} & 5 & Min.\ face quality to keep frame in memory \\
            & \texttt{MIN\_SETTING\_QUALITY} & 5 & Min.\ setting quality to keep frame in memory \\
            & \texttt{NMS\_WINDOW} & 5 & Temporal NMS window (sec) \\
            & \texttt{MIN\_SCORED\_FRAMES} & 3 & Min.\ number of frames kept even for very short clips \\
            & \texttt{FACE\_CLUSTER\_THRESHOLD} & 0.45 & ArcFace cosine similarity threshold for character clustering \\
            & \texttt{CLUSTER\_VERIFY\_THRESHOLD} & 0.40 & ArcFace cosine similarity threshold to retain faces in cluster \\
            & \texttt{MIN\_FACE\_DET\_SCORE} & 0.5 & Min.\ InsightFace detection confidence for cluster pruning \\
            & \texttt{MIN\_FACE\_CLIPS} & 2 & Min.\ number of clips a character must appear in for cluster pruning \\
            & \texttt{MIN\_IDENTITY\_FACE\_Q} & 7 & Min.\ face quality to index frame as character reference \\
        \midrule
        \multirow{3}{*}{Retrieval}
            & MAX\_CHARACTER\_REFS & 4 & Char. ref. images per shot \\
            & MAX\_SETTING\_REFS & 2 & Setting ref. images per shot \\
            & MIN\_SIMILARITY & 0.35 & Min. cosine similarity for retrieved refs. \\
        \midrule
        \multirow{8}{*}{Generation}
            & MIN\_TRIM\_DURATION & 1 & Min. clip length after trimming (sec) \\
            & BEST\_OF\_N & 3 & Keyframe candidates generated per shot \\
            & CRITIC\_MAX\_ITERS & 2 & Max. critic rounds \\
            & MIN\_KEYFRAME\_SCORE & 7 & Retry if best-of-N score below this \\
            & YAW\_THRESHOLD & 60 & ArcFace unreliable above this yaw; fall back to VLM (deg) \\
            & ArcFace weight ramp lo/hi & 0.01/0.10 & Face-fraction thresholds for ArcFace weight ramp \\
            & $w_c$ & [0.50, 0.85] & ArcFace weight applied to score \\
            & $w_{type}$ & [0.15, 0.95] & Char. vs. setting weight by shot type \\
        \bottomrule
    \end{tabular}
\end{table*}

\subsection{Prompts}

All LLM/VLM prompts can be found in Figures \ref{fig:clip-analysis-prompt}-\ref{fig:set-judge-prompt}.

\subsection{Longer B-Roll Performance}

We isolate the user study responses for 7 of the 39 evaluation prompts with $>$30 second target durations (max 53s) in \Cref{tab:longer-baselines,tab:longer-ablation}, and find that the trends roughly follow the results in \Cref{tab:baselines,tab:ablation}.

\begin{table}[H]
    \caption{Comparison against baselines from A/B user studies for longer output sequences ($>$30s).
    The green bar and number indicate the win rate for our approach, and the red bar indicates the win rate for the baseline. * indicates a p-value $<0.02$.}
    \label{tab:longer-baselines}
    \resizebox{\linewidth}{!}{%
    \begin{tabular}{cccc}
        \toprule
        Method & $|\Delta T|$ (s) $\downarrow$ & Prompt adherence ($\mathcal{U}$) & Visual alignment ($\mathcal{V}$)\\
        \midrule
        T2V + Planner & 1.26 $\pm$ 0.95 & \prefbar{34.7}* & \prefbar{8.5}* \\
        EditDuet & 4.77 $\pm$ 4.75 & \prefbar{4.3}* & \prefbar{66.2}* \\
        \method & \textbf{0.12 $\pm$ 0.05} & N/A & N/A \\
        \bottomrule
    \end{tabular}
    }
\end{table}

\begin{table}[H]
    \caption{Ablation study results from A/B user studies for longer output sequences ($>$30s).
    The green bar and number indicate the win rate for our full approach, and the red bar indicates the win rate for the row variant. * indicates a p-value $<0.02$.}
    \label{tab:longer-ablation}
    \resizebox{\linewidth}{!}{%
    \begin{tabular}{cccc}
        \toprule
        Stage & Variant & Prompt adherence ($\mathcal{U}$) & Visual alignment ($\mathcal{V}$)\\
        \midrule
        \multirow{3}{*}{Memory}
        & Uniform & \prefbar{53.2}\phantom{*} & \prefbar{15.6}* \\
        & Scene-detected & \prefbar{42.9}\phantom{*} & \prefbar{11.5}* \\
        & Uniform + C & \prefbar{37.8}\phantom{*} & \prefbar{41.4}\phantom{*} \\

        \midrule
        \multirow{3}{*}{Retrieval}
        & Random & \prefbar{28.6}* & \prefbar{48.0}\phantom{*} \\
        & Caption-RAG & \prefbar{41.7}\phantom{*} & \prefbar{35.9}\phantom{*} \\
        & w/o char. ID gate & \prefbar{40.0}\phantom{*} & \prefbar{31.6}\phantom{*} \\

        \midrule
        \multirow{2}{*}{Generation}
        & w/o judges & \prefbar{37.7}\phantom{*} & \prefbar{38.3}\phantom{*} \\
        & w/o critic & \prefbar{55.4}\phantom{*} & \prefbar{41.6}\phantom{*} \\
        \bottomrule
    \end{tabular}
    }
\end{table}

\begin{figure*}
\begin{lstlisting}[basicstyle=\small\ttfamily, breaklines=true, frame=single]
Analyze this video. Extract characters, props, location, actions. Be thorough.
For each character, strictly separate permanent vs. transient traits:
- physical_description: ONLY permanent traits (ethnicity, gender, age, build,
hair color/texture, facial features, facial hair, scars, tattoos)
- appearance: everything that could change between clips (clothing, hairstyle,
accessories, glasses, hats, jewelry, makeup, held props)
\end{lstlisting}
\caption{VLM prompt for clip-level analysis (native video upload).}
\label{fig:clip-analysis-prompt}
\end{figure*}

\begin{figure*}
\begin{lstlisting}[basicstyle=\small\ttfamily, breaklines=true, frame=single]
Analyze this frame for reference quality.
- setting_quality: 1-10 (10=clear wide view of environment with good lighting
  and composition, shows key details of the space; 1=too close-up or blurry
  to understand the setting)
- view_type: one of ['frontal', 'three_quarter', 'profile', 'full_body',
  'hands_detail', 'back', 'over_shoulder', 'close_up_detail'] (overall frame composition)
- view_description: brief label of what is visible
  (e.g. "wide view of rustic bakery kitchen", "medium shot of two people at a table")
- faces: an array of EVERY individual face visible in the frame. For each face return:
  - face_quality: 1-10 (10=perfect clear face, 1=barely visible/blurry)
  - view_type: one of ['frontal', 'three_quarter', 'profile', 'back']
  - description: brief physical description of THIS person
    (e.g. "woman with short grey hair and glasses", "young man with dark beard")
If no faces are visible, return an empty array.
Return JSON: {"setting_quality": N, "view_type": "...", "view_description": "...",
"faces": [{"face_quality": N, "view_type": "...", "description": "..."}, ...]}
Return ONLY valid JSON.
\end{lstlisting}
\caption{VLM prompt for frame scoring (per frame).}
\label{fig:frame-scoring-prompt}
\end{figure*}

\begin{figure*}
\begin{lstlisting}[basicstyle=\small\ttfamily, breaklines=true, frame=single]
These images all show the same person from different clips.
Write a detailed physical description: ethnicity, gender, approximate age,
hair style/color, facial features, glasses, facial hair, build.
Focus on PERMANENT features (not clothing, since that changes across clips).
Be specific enough that this description could uniquely identify them.
Return ONLY the description, one paragraph.
\end{lstlisting}
\caption{VLM prompt for character sheet descriptions (up to 4 best frames).}
\label{fig:char-sheet-prompt}
\end{figure*}

\begin{figure*}
\begin{lstlisting}[basicstyle=\small\ttfamily, breaklines=true, frame=single]
Summarize this video collection for a b-roll planner that needs to write
retrieval queries against it. Be specific about what's available.

Clip descriptions ({N} clips):
{desc_block}

Produce a compact inventory covering:
1. Settings/locations: what environments exist, rough clip count each
2. Activities/actions: what people are doing across the collection
3. Props/objects: notable items that appear repeatedly

Do NOT include a Characters section -- character descriptions are stored
separately in character_sheets.json and keyed by character_id; duplicating
them here with ad-hoc labels ('Man in Maroon') risks the planner attaching
one character's wardrobe to another's beats.

Keep it under 500 words. Use specific visual details that would help write
good retrieval queries. Group similar clips together.
\end{lstlisting}
\caption{LLM prompt for collection summary (all clip descriptions).}
\label{fig:collection-summary-prompt}
\end{figure*}

\begin{figure*}
\begin{lstlisting}[basicstyle=\small\ttfamily, breaklines=true, frame=single]
You write concise, single-sentence visual descriptions of images.
Follow the scope instructions exactly -- include only the categories named
under INCLUDE, exclude anything under EXCLUDE. Present tense, no hedging,
no preamble.

Describe this person's PERMANENT identity features.
INCLUDE: ethnicity, gender, approximate age, face structure, hair color and
general style, distinctive permanent features (glasses frames, facial hair).
EXCLUDE: outfit, clothing, jewelry, accessories, or anything that could
change between scenes.
One sentence.

Describe this person's TRANSIENT attributes.
INCLUDE: current outfit (shirt, apron, pants, shoes), visible jewelry,
accessories, hair styling state.
EXCLUDE: ethnicity, gender, age, face structure, or hair color (handled
separately); the environment or any background.
One sentence.
\end{lstlisting}
\caption{VLM prompt for permanent and transient character attribute extraction.}
\label{fig:char-attr-prompt}
\end{figure*}

\begin{figure*}
\begin{lstlisting}[basicstyle=\small\ttfamily, breaklines=true, frame=single]
You write concise, single-sentence visual descriptions of images.
Follow the scope instructions exactly -- include only the categories named
under INCLUDE, exclude anything under EXCLUDE. Present tense, no hedging,
no preamble.

Describe the PERMANENT features of this environment.
INCLUDE: location type, architectural features, fixed equipment and furniture,
dominant color palette.
EXCLUDE: transient props on surfaces, current lighting state, or arrangement
details that could change.
One sentence.

Describe this environment's TRANSIENT state.
INCLUDE: current props on surfaces, lighting quality and time-of-day, what is
open or closed, current arrangement of movable items.
EXCLUDE: the location type, architectural features, or fixed equipment
(handled separately); any people in the frame (character state is handled
separately).
One sentence.
\end{lstlisting}
\caption{VLM prompt for permanent and transient setting attribute extraction.}
\label{fig:setting-attr-prompt}
\end{figure*}

\begin{figure*}
\begin{lstlisting}[basicstyle=\small\ttfamily, breaklines=true, frame=single]
You track transient scene state across shots of a b-roll sequence.
Produce clean present-tense prose describing the current state; never
annotate changes or reference prior descriptions (no 'now wearing',
'replacing', 'updated from', 'previously'). The reader should see a
self-contained description.

[Image 1: img]  [Image 2: img]  ...

Focus EXCLUSIVELY on {scope_focus}.
Exclude: {scope_exclude}.

Produce ONE coherent description capturing every transient attribute visible
across these images. When the same attribute appears with different values,
pick the MAJORITY version (or the latest on ties) for the description and list
the contradiction in `conflicts` as a short phrase (e.g. 'shirt color: white
in 2, blue in 1'). Output must be a SINGLE paragraph -- never include two
sentences that describe the same attribute, and never preserve a previous
paragraph verbatim alongside a rephrased version.

[if character_focus set:]
CHARACTER FOCUS: describe ONLY the person matching this permanent description
(use stable identifiers like ethnicity, build, hair, beard, glasses to
identify them):
"{character_focus}"
If multiple people are visible, ignore everyone who clearly doesn't match.
If no one in the image matches, return an empty description and an empty
conflicts list.

[if prev_merged set:]
Previous merged description (preserve attributes unless the new images
clearly contradict them):
"{prev_merged}"

[if state_updates set:]
Staged scene-evolution updates to integrate (these supersede matching
attributes in the previous description; merge seamlessly -- do NOT copy the
raw wording or leave a duplicate paragraph):
"{state_updates}"
\end{lstlisting}
\caption{VLM prompt for transient state tracking.}
\label{fig:transient-tracking-prompt}
\end{figure*}

\begin{figure*}
\begin{lstlisting}[basicstyle=\small\ttfamily, breaklines=true, frame=single]
You are a b-roll narrative planner. You decide WHAT the viewer sees --
beats, characters, settings, emotional arc -- not HOW to shoot it
(shot types, motion, precise durations are assigned downstream).

B-roll is supplementary footage: no talking heads, no direct address to
camera, no interviews. People doing or interacting is fine. People
looking into the lens and speaking is not.

NARRATIVE RULES (for planning story beats):

CUT PRIORITY (Walter Murch's Rule of Six -- sacrifice from the bottom up):
  1. Emotion (51%) -- true to the feeling of the moment
  2. Story (23%) -- advances the narrative
  3. Rhythm (10%) -- occurs at a rhythmically right moment
  When rules below conflict with emotion or story, emotion wins.

SEQUENCE STRUCTURE:
- Open with an establishing beat to set context (who, where, when).
- Close with a resolving beat for emotional resolution.
- Every beat should deliver new information or shift emotion -- if neither,
  the beat is unmotivated and should be removed.
- Begin and end moments with action in progress -- avoid a staged feeling.

SCREEN DIRECTION:
- Maintain consistent spatial relationships for the same subject across beats.

CONTENT VARIETY:
- Each beat should show something meaningfully different from its neighbors --
  a new angle, new detail, new action, or new emotional register.
- Avoid repeating the same type of moment (e.g. three consecutive "person
  walks" beats).

SPECIFICITY:
- Mirror the user's level of specificity in setting_query, identity_query,
  and generation_prompt. Add visual concreteness (texture, lighting, scale,
  materials) where it helps retrieval, but do NOT introduce cultural,
  architectural, geographic, or temporal specifics that aren't in the user's
  words. The collection summary is context for what's available -- not a menu
  of attributes to graft onto the user's intent.
\end{lstlisting}
\caption{LLM system instruction for the narrative planner.}
\label{fig:narr-planner-instruction}
\end{figure*}

\begin{figure*}
\begin{lstlisting}[basicstyle=\small\ttfamily, breaklines=true, frame=single]
Prompt: "{prompt}"
Target duration: ~{duration}s total; each beat becomes a 1-8s clip.

Collection sample:
{collection_context}

[if character_sheets:]
Known characters in the collection:
  - character_00 ({N} clips): {description}
  - character_01 ({N} clips): {description}

Assigning character_ids to each beat:
- If the prompt names a specific character (e.g. character_00), include
  that character_id.
- If the prompt describes a person by role or appearance without naming them
  (e.g. "the man", "a teacher", "a violinist"), pick the character_id from
  the list above whose description best matches the role/context. This keeps
  identity refs consistent across shots of the same person.
- Only leave character_ids empty [] when (a) the beat contains no person
  at all, or (b) no character in the list plausibly matches the described role.

Emit beats with these fields:
- character_ids: list of character_ids featured (see character context above).
- identity_queries: list of per-character retrieval queries. Emit ONE entry
  per character_id in this beat: [{"character_id": "character_00", "query":
  "..."}, ...]. Each query is a natural-language description of THAT character
  alone.
  Each query should focus on STABLE identifiers (ethnicity, build, face, hair,
  always-worn items like prescription glasses).
  Empty list when needs_character is false. Always emit identity_queries when
  needs_character is true and character_ids is non-empty.
- setting_query: retrieval query for the environment.
- generation_prompt: scene description for image generation (subject + action
  + environment). Skip wardrobe unless the user prompt asks for it or the beat
  needs an outfit change.
- needs_character: bool.
- setting_in_collection: bool. Does the "Collection sample" above plausibly
  contain footage of this beat's setting/environment? TRUE if the intended
  location appears in the collection (even with different framing/angle).
  FALSE if the setting is genuinely absent (e.g. user prompt asks for an
  indoor pool but the collection only lists outdoor lakes). When FALSE, the
  highest-scoring collection frame is used as a fallback reference. Be honest
  -- marginal matches that only kind of work should be FALSE. When in doubt
  for ambient/generic settings, lean TRUE.
- character_in_collection: bool. When needs_character is TRUE, does the
  assigned character_ids list point to someone actually in the collection?
  Usually TRUE when character_ids is non-empty. Set FALSE only when
  needs_character is TRUE but no character_id was assignable. Set TRUE when
  needs_character is FALSE (field is ignored).
- beat_duration_hint: 1-8s; hints roughly sum to {duration}s (exact durations
  assigned later).

Self-check before returning: logical progression, emotional arc, each beat
meaningfully distinct, delivers the prompt.

SHOT COUNT OVERRIDE: if the prompt specifies an exact number of shots/beats
("single shot", "two shots", "3 shots"), produce EXACTLY that many. No
establishing/closer additions. Overrides all cinematic rules.
\end{lstlisting}
\caption{LLM prompt for the narrative planner.}
\label{fig:narr-planner-prompt}
\end{figure*}

\begin{figure*}
\begin{lstlisting}[basicstyle=\small\ttfamily, breaklines=true, frame=single]
You are the assembly editor for a b-roll sequence. Given narrative beats,
you assign cinematic execution: shot_type, motion_type, conditioning_mode,
video_prompt, and exact trim_duration.

CINEMATIC RULES (apply to all sequences):

CUT PRIORITY (Walter Murch's Rule of Six -- sacrifice from the bottom up):
  1. Emotion (51%) -- true to the feeling of the moment
  2. Story (23%) -- advances the narrative
  3. Rhythm (10%) -- occurs at a rhythmically right moment
  4. Eye-trace (7%) -- respects where the viewer is looking
  5. Composition (5%) -- respects 2D screen grammar
  6. Spatial continuity (4%) -- consistent screen direction

SHOT SCALE:
- Consecutive shots of the SAME SUBJECT must differ by >=2 steps on the scale
  ladder (ECU > CU > MCU > MS > MLS > LS > ELS). 1 step risks a jump cut;
  >3 steps is a disorienting spatial jump. Shots of DIFFERENT subjects can
  jump freely.
- Scale-in (wider -> tighter) feels natural and draws the viewer in. Scale-out
  (tighter -> wider) is more abrupt -- use it deliberately for reveals.
- Angle shots (low_angle, high_angle, dutch_angle, aerial, POV,
  over_the_shoulder) are wildcards: they pair with any scale and reset the
  ladder for the next shot.

CAMERA MOTION:
- Static-buffer rule: do NOT cut directly from one camera move to another.
  Insert a static shot between two moving shots as a visual reset. Exception:
  whip-pan-to-whip-pan in the SAME direction at matching speed.
- No direction reversals on consecutive shots of the same subject (e.g.
  pan_left then pan_right). Use a static shot or change subject between them.
- Match motion energy across cuts: moving -> moving or static -> static. Avoid
  fast motion directly to locked-off static unless the contrast is intentional.

DURATION & PACING:
- Vary clip durations -- uniform lengths feel mechanical. Never use the same
  duration for more than two consecutive shots.
- Static shots should generally be shorter than moving shots (less motion =
  information is delivered faster; movement continuously reveals new content).
- Wide shots need more time (more visual information to absorb). Close-ups can
  be shorter (focused attention, quick read).
- Pacing should have peaks and valleys -- accelerate for energy, decelerate for
  contemplation. A metronome is not editing.

SEQUENCE STRUCTURE:
- Open with a wide establishing shot (ELS or LS) to set context, then
  progressively tighten into the subject/detail (the classic establishing
  sequence).
- Close with a resolving beat: either pull back to a wide shot (bookend), or
  hold the final shot longest for emotional resolution.
- Every cut should deliver new information or shift emotion -- if neither, the
  cut is unmotivated and should be removed.
- Begin and end shots with action in progress -- avoid the "staged" feeling of
  subjects entering or exiting frame from nothing.

SCREEN DIRECTION:
- Maintain consistent screen direction for the same subject across shots (if
  they face right in shot 1, they face right in shot 2) unless deliberately
  reversing for narrative effect. Insert a neutral cutaway to cross the axis
  cleanly.
\end{lstlisting}
\caption{LLM system instruction for assembly planner.}
\label{fig:asmb-planner-instruction}
\end{figure*}

\begin{figure*}
\begin{lstlisting}[basicstyle=\small\ttfamily, breaklines=true, frame=single]
Original prompt (context): "{prompt}"
Target total: exactly {duration}s. Number of beats: {N} -- produce EXACTLY one
shot per beat.

Editing style rules:
{style_rules}

Narrative beats:
{beats_json}

For each beat assign:
- shot_type: one of [extreme_close_up, close_up, medium_close_up, medium_shot,
  medium_long_shot, long_shot, extreme_long_shot, over_the_shoulder,
  point_of_view, aerial, low_angle, high_angle, dutch_angle]
- motion_type: one of [static, pan_left, pan_right, tilt_up, tilt_down,
  dolly_in, dolly_out, tracking, crane_up, crane_down, orbit, handheld,
  zoom_in, zoom_out]
- conditioning_mode: "first_frame" (default) or "reference". Use "reference"
  when the clip reveals angles/content not in the keyframe (subject turns,
  rotates, exposes new side). orbit/tracking/crane/pan auto-force "reference".
- video_prompt: action + environment + style. Use force-based verbs (press,
  slide, pour, lift), concrete sensory details, lighting/palette. Exclude
  camera movement (handled by motion_prompt) and subject appearance (handled
  by refs). No negative constraints. SINGLE continuous moment -- no transition
  language ("fade in", "cut to", "then zoom to").
  TONE: write naturalistic, observational descriptions. Veo amplifies --
  understate the action. Avoid hyperbolic intensifiers ("explosive", "massive",
  "huge", "epic", "intense", "powerful", "dramatic") and atmospheric
  exaggerations (large dust clouds, sparks, motion blur, slow-motion drama)
  unless the user prompt explicitly asks for them. Force-based verbs (press,
  slide, lift) are fine; aggressive ones (slam, smash, charge, leap) tend to
  over-trigger Veo into stylized spectacle.
- trim_duration: 1-8s decimals. ALL trim_durations MUST sum to exactly {duration}.

USER-INTENT OVERRIDE: if the original prompt explicitly specifies shot scale,
motion type, or camera behavior (e.g. "static", "handheld camera", "extreme
close-up", "wide shots only", "slow-motion"), treat those as hard constraints
that override the style rules above for ALL shots.

Produce exactly {N} shot(s).
\end{lstlisting}
\caption{LLM prompt for the assembly planner.}
\label{fig:asmb-planner-prompt}
\end{figure*}

\begin{figure*}
\begin{lstlisting}[basicstyle=\small\ttfamily, breaklines=true, frame=single]
You are a b-roll sequence critic. You evaluate keyframes against an
10-dimension rubric, flag issues, and emit structured shot actions
(keep / regenerate / remove / retime) plus optional new_shots.

DURATION BUDGET -- TREAT AS A FIXED-SUM TRANSACTION.
The final sequence must hit the target duration exactly. Time is a closed
budget: you cannot 'add' time, you can only redistribute it.
Every ADD (new_shot) must be paid for by an offsetting REMOVE or RETIME on
existing shots. Every RETIME that increases a duration must be balanced by a
decrease elsewhere. The sum of `delta_seconds` across all entries in
`duration_ledger` MUST equal 0.
Before deciding on actions, compute the budget gap (current vs target). If
the gap is small, prefer RETIME over ADD/REMOVE. If you must ADD, explicitly
pair it with a REMOVE or shrink-RETIME of equal magnitude.

CINEMATIC RULES (apply to all sequences):
[... same as SYSTEM_ASSEMBLY_PLANNER cinematic rules above ...]

## EVALUATION RUBRIC
Evaluate each dimension below. For each, state PASS or FAIL with a brief reason.

ACTIONABILITY REQUIREMENT: every FAIL must be addressed by at least one
concrete shot_action (regenerate / retime / remove / new_shot). Do not leave
a FAIL without a corresponding action. In the reason field, reference the
specific shot_action(s) that will fix it. If a FAIL genuinely cannot be fixed
with the available actions, explain why -- but this should be rare. Unaddressed
FAILs are treated as unresolved issues that will trigger another iteration.

USER-INTENT OVERRIDE: when the user prompt explicitly specifies a framing or
pacing constraint (e.g. "all tight close-ups", "only handheld", "no wide
shots", "matched-scale only", "single shot", "uniform pacing"), treat that as
a hard constraint and do NOT FAIL dimensions (4a) shot scale, (4b)
static-buffer, (4c) motion direction, or (4d) sequence structure for honoring
the user's explicit framing. Cite the prompt phrase that licensed the override
in your reason. Identity & setting scores (3), visual consistency (5),
motivated cuts (4e), and screen direction (4f) still apply unconditionally.

(1) CONTENT ALIGNMENT: Does each shot serve the original prompt?

(2) VISUAL VARIETY: Are motion types and angle types sufficiently varied?
Flag repetition (e.g. 3+ consecutive shots with the same motion, or 3+
consecutive shots from the same angle). Shot-scale variety is governed by (4a)
and is subject to the USER-INTENT OVERRIDE -- do NOT re-flag scale repetition
here.

(3) IDENTITY & SETTING SCORES: Flag shots scoring below 6 on either dimension
as clear failures. Shots scoring 6.0-6.9 are borderline -- only flag for
`regenerate` when there's a specific, plausible fix. Avoid generic "low score"
regens on borderline shots, since the keyframe retry loop has already tried
multiple attempts; another blind regen rarely helps and consumes the iteration
budget.
\end{lstlisting}
\caption{VLM system instruction for the critic (part 1).}
\label{fig:critic-instruction-1}
\end{figure*}

\begin{figure*}
\begin{lstlisting}[basicstyle=\small\ttfamily, breaklines=true, frame=single]
(4) CINEMATIC GRAMMAR -- check each rule for every adjacent pair of shots.
SHOT-COUNT GATE: skip 4a-4d entirely when the sequence has fewer than 3
shots -- adjacency-based grammar and sequence-structure checks are vacuous or
impossible at that scale. (4e) motivated cuts and (4f) screen direction still
apply at any shot count.
  4a. SHOT SCALE: Do consecutive same-subject shots differ by >=2 steps on
      the ladder? (Flag jump cuts from insufficient scale change.)
  4b. STATIC-BUFFER: Is there a static shot between any two moving-camera
      shots? (Flag movement-to-movement cuts unless the style explicitly
      overrides this.)
  4c. MOTION DIRECTION: Are there direction reversals on consecutive
      same-subject shots (e.g. pan_left -> pan_right without a static reset)?
  4d. SEQUENCE STRUCTURE: Does the sequence open with an establishing wide
      shot? Does it close with a resolving beat (longest shot or deliberate
      bookend)?
  4e. MOTIVATED CUTS: Does every shot deliver new information or shift
      emotion? Flag any shot that feels redundant.
  4f. SCREEN DIRECTION: Does the same subject maintain consistent screen
      direction?

(5) VISUAL CONSISTENCY -- this is critical:
  5a. Character identity coherence: the SAME person must look like the same
      person across ALL shots. Focus on PERMANENT attributes: face structure,
      skin tone, ethnicity, gender, build, and age.
  5b. Transient attribute coherence: clothing, jewelry, accessories, and
      hairstyle MUST be consistent across shots in the same scene. Flag any
      mismatch for regeneration.
  5c. Lighting/time-of-day consistency across shots in the same scene.
  5d. Spatial coherence of background elements across shots in the same scene.
\end{lstlisting}
\caption{VLM system instruction for the critic (part 2).}
\label{fig:critic-instruction-2}
\end{figure*}

\begin{figure*}
\begin{lstlisting}[basicstyle=\small\ttfamily, breaklines=true, frame=single]
Keyframe stills (in shot order):
[img] [img] ...

Original prompt: "{prompt}"
Failed: {n_failed}/{total}. Iteration: {iteration+1}.

Duration budget (per-shot current durations):
  shot 0: X.XXs
  shot 1: X.XXs
  ...
  TOTAL: XX.XXs  TARGET: {duration}s  GAP: +/-X.XXs

Style: {style_key}
{style_rules}

Shots:
i=0  medium_shot         static        6.00s  id= 8.2  set= 9.1  veo=first_frame   OK
i=1  close_up            dolly_in      4.00s  id= 7.5  set= 8.8  veo=first_frame   OK  2 conflicts
...

[if conflicts:]
Conflicts flagged (cross-shot drift -- prioritize regeneration; pick the
shot(s) inconsistent with the sequence):
  i=1: id: shirt color: white in shot 0, grey in shot 1

[if identity details:]
Identity attribute breakdown (per-shot VLM + ArcFace; use these to name the
failing attribute in your regen reasons):
  i=1: mode=hybrid, arcface=8.3, clothing=3, accessories=9, face_plausibility=9

Shot intents:
  i=0: "{generation_prompt[:120]}"
  i=1: "{generation_prompt[:120]}"

`trim_duration` on every shot_action is the FINAL duration for that shot
(not a delta). trim_duration range: 1-8s (decimals OK).

Duration arithmetic: emit `duration_ledger` with one entry per action that
affects time. Use shot_idx=-1 for new_shots and current=0; for REMOVE set
proposed=0 (delta=-current); for KEEP/REGENERATE/RETIME use the existing
index. sum(delta_seconds) MUST equal {gap:+.2f}s (target {duration}s -
current {planned}s). Also put a one-line summary in `duration_math`.
\end{lstlisting}
\caption{VLM prompt for the sequence-level critic.}
\label{fig:critic-prompt}
\end{figure*}

\begin{figure*}
\begin{lstlisting}[basicstyle=\small\ttfamily, breaklines=true, frame=single]
You compare a reference and a generated image for identity consistency.
Score 1-10 per dimension. Ignore lighting and pose differences; focus on
whether it is the same person and whether wardrobe/accessories match.

Reference: [img]  Generated: [img]
Score each dimension 1-10:
- clothing_fidelity: same outfit (color, style, layers, patterns)?
- accessories_fidelity: same glasses, jewelry, hat, watch, bag?
- face_plausibility: could this be the same person? 8-10 clearly yes,
  4-6 uncertain, 1-3 obviously different (wrong ethnicity/gender/age).
- overall_attribute_score: weighted toward transient attributes but penalized
  if face is obviously wrong.
\end{lstlisting}
\caption{VLM prompt for the character judge.}
\label{fig:char-judge-prompt}
\end{figure*}

\begin{figure*}
\begin{lstlisting}[basicstyle=\small\ttfamily, breaklines=true, frame=single]
You compare a generated image against setting references for environment
fidelity: location type, lighting, color palette, props, atmosphere.
Ignore the people in the frame; score only the environment.

Setting reference: [img]
[Setting reference: [img]]
Generated image: [img]
setting_fidelity (1-10). 10=exact same location. 7-8=same type of place with
similar feel. 5-6=vaguely similar. 3-4=different place. 1-2=completely wrong
environment. Consider location type, lighting, color palette, key
props/furniture, atmosphere.
\end{lstlisting}
\caption{VLM prompt for the setting judge.}
\label{fig:set-judge-prompt}
\end{figure*}

\end{document}